\pdfoutput=1

\documentclass[11pt]{article}

\usepackage[final]{acl}
\usepackage{booktabs}
\usepackage{easyReview}
\usepackage{times}
\usepackage{latexsym}
\usepackage{CJKutf8}
\usepackage{enumitem}
\usepackage[T1]{fontenc}

\usepackage{circledsteps}

\usepackage[utf8]{inputenc}
\usepackage{url}
\usepackage{microtype}

\usepackage{inconsolata}

\usepackage{graphicx}
\usepackage{multirow}
\usepackage{amsmath}
\usepackage{graphicx}
\usepackage{booktabs}
\newcommand{\ours}{\textsc{Ress}}
\newcommand{\oursdata}{\textsc{Cheer}}
\title{Can Large Language Models Understand Internet Buzzwords \\Through User-Generated Content }

\author{
Chen Huang$^{\spadesuit\clubsuit}$, \quad
Junkai Luo$^{\spadesuit}$, \quad
Xinzuo Wang$^{\diamondsuit}$, \quad
Wenqiang Lei$^{\spadesuit\clubsuit}$\thanks{Correspondence to Wenqiang Lei.}, \quad
Jiancheng Lv$^{\spadesuit\clubsuit}$
\\
$^{\spadesuit}$College of Computer Science, Sichuan University, China \\
$^{\clubsuit}$Engineering Research Center of Machine Learning and Industry Intelligence, \\Ministry of Education, China \quad
$^{\diamondsuit}$JD.com, China \\
\texttt{huangc.scu@gmail.com, luojunkai@stu.scu.edu.cn, wangxinzuo@jd.com} \\ \texttt{\{wenqianglei, lvjiancheng\}@scu.edu.cn}
}

\begin{document}
\maketitle
\begin{abstract}
The massive user-generated content (UGC) available in Chinese social media is giving rise to the possibility of studying internet buzzwords. In this paper, we study if large language models (LLMs) can generate accurate definitions for these buzzwords based on UGC as examples. Our work serves a threefold contribution. First, we introduce \oursdata, the first dataset of Chinese internet buzzwords, each annotated with a definition and relevant UGC. Second, we propose a novel method, called \ours, to effectively steer the comprehending process of LLMs to produce more accurate buzzword definitions, mirroring the skills of human language learning. Third, with \oursdata, we benchmark the strengths and weaknesses of various off-the-shelf definition generation methods and our \ours. Our benchmark demonstrates the effectiveness of \ours~while revealing crucial shared challenges: over-reliance on prior exposure, underdeveloped inferential abilities, and difficulty identifying high-quality UGC to facilitate comprehension. We believe our work lays the groundwork for future advancements in LLM-based definition generation. Our dataset and code are available at \url{https://github.com/SCUNLP/Buzzword}.
\end{abstract}

\section{Introduction}
\begin{figure}
    \centering
    \setlength{\abovecaptionskip}{2pt}   
    \setlength{\belowcaptionskip}{2pt}
    \includegraphics[width=0.5\textwidth]{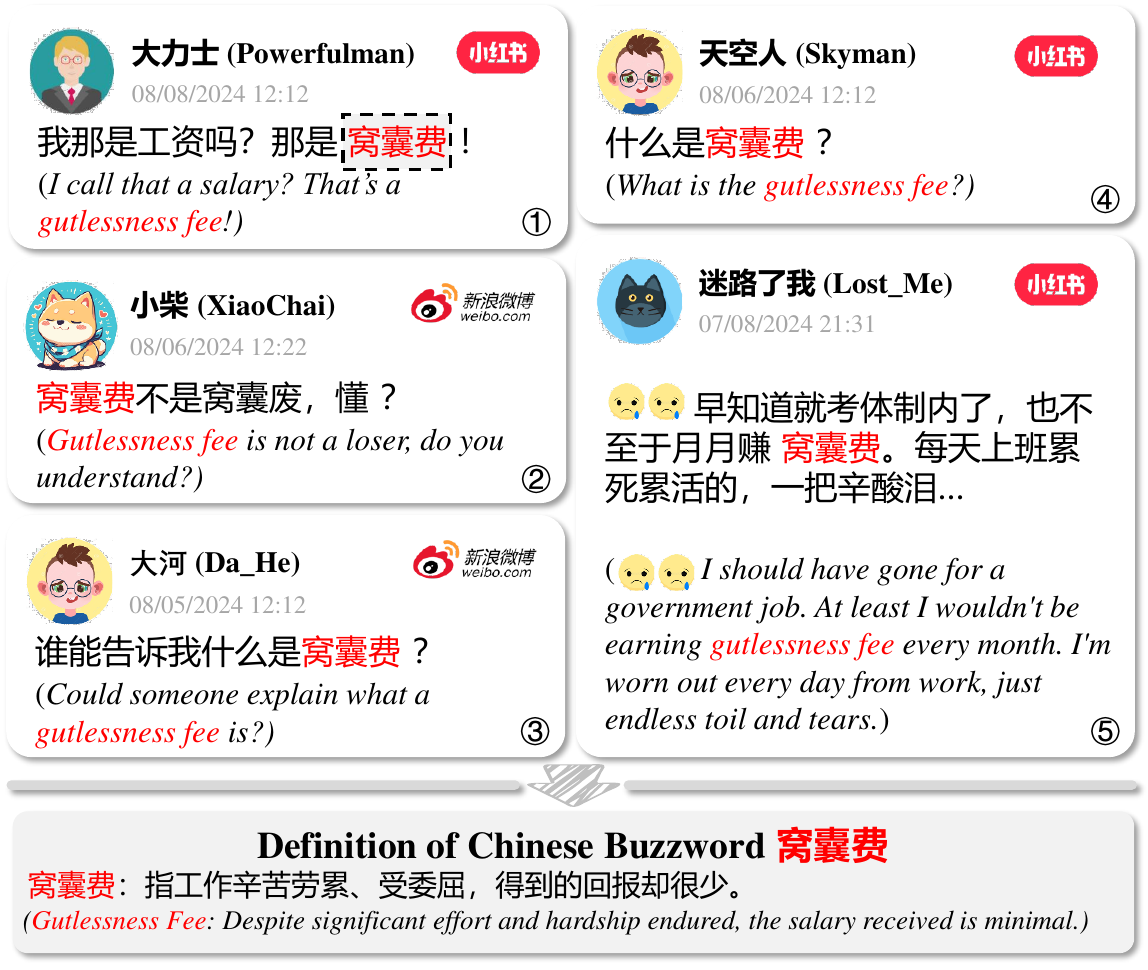}
    \caption{Task Illustration: generating definitions for Chinese buzzwords using UGC.}
    \label{fig:illus}
    \vspace{-5mm}
\end{figure}

Internet buzzwords emerge as newly coined terms representing abstract concepts that may extend beyond their literal definitions \cite{malyuga2021making}, such as `\begin{CJK}{UTF8}{gbsn}窝囊费\end{CJK}'(i.e., \textit{gutlessness fee}). They often rapidly gain popularity through social media platforms and are amplified by user-generated content (UGC) such as posts and reviews \cite{liu2020reliable}. 
However, their inherent abstractness creates ambiguity, presenting significant challenges for human understanding without further explanation \cite{tsur2015don, cornwall2007buzzwords, malyuga2021making, huang-etal-2022-understanding}. For instance, as illustrated in Figure \ref{fig:illus}, the Chinese buzzword `\begin{CJK}{UTF8}{gbsn}窝囊费\end{CJK}' is commonly used to reflect user dissatisfaction with inadequate compensation for their efforts, highlighting a feeling of being undervalued rather than the literal meaning of `\begin{CJK}{UTF8}{gbsn}窝囊\end{CJK}', which implies cowardice. 
Given the absence of these terms in traditional dictionaries, online UGC becomes a crucial resource for understanding their meaning. Therefore, \textbf{our task, as shown in Figure \ref{fig:illus}, is to generate definitions for internet buzzwords using UGC as illustrative examples}, bridging the gap between their widespread usage and a precise understanding of their meaning.

Formally, our task falls under the category of context-aware definition generation \cite{li2020explicit, huang2021definition}, which involves automatically generating dictionary definitions by incorporating both the target word and contextual information, such as example sentences. While existing methods effectively learn to define common words using static training data \cite{mickus2019mark, zheng2021decompose,  huang2021definition, zhang2023assisting}, they struggle with the dynamic nature of online buzzwords, which emerge and disappear rapidly.
Even advanced large language models (LLMs) exhibit limitations with long-tail words \cite{wu2024can, rodriguez2023exploring}, let alone newly coined buzzwords. Therefore, \textbf{effectively comprehending buzzwords remains a significant challenge}.

To substantiate our analysis, we construct a dataset of \underline{CH}in\underline{E}s\underline{E} internet buzzwo\underline{R}ds, called \textbf{\oursdata}, and evaluate the performance of existing methods (cf. Section \ref{benchmark}). The evaluation results clearly demonstrate that existing methods, including LLM-based ones, struggle to accurately define buzzwords.
Therefore, leveraging UGC to imbue LLMs with an understanding of rapidly evolving buzzwords poses a significant challenge. 

To tackle the challenge, we propose \textbf{\ours}~to stee\underline{R} the compr\underline{E}hending proce\underline{SS} of LLMs to produce buzzword definitions. By mimicking human language acquisition, \ours~codifies key comprehension skills into distinct aspects, prompting the LLM to produce aspect-specific definition candidates. These candidates are then integrated through an ensemble process for a more nuanced and accurate final definition, enriching the LLM's understanding towards internet buzzwords.

To thoroughly evaluate the efficacy and limitations of existing methods and \ours~across various LLM backbones, we establish a benchmark using our curated dataset \oursdata. Our results demonstrate that \ours~surpasses baseline performance, achieving an average improvement of +2.51\% in semantic accuracy and +3.31\% in semantic completeness\footnote{Semantic completeness refers to a definition encompassing all and only the relevant aspects of a word's meaning.} over the best baseline. Despite these gains, the overall performance of all methods remains suboptimal. Further analysis reveals key challenges LLMs face in interpreting buzzwords derived from UGC. Specifically, we identify an overreliance on prior exposure to buzzwords\footnote{Performance is significantly reduced on unseen buzzwords compared to seen ones.} and a limited capacity to infer the meaning of buzzwords, hindering their ability to handle the dynamic nature of internet buzzwords. Additionally, both the volume and quality of UGC are critical factors influencing definition accuracy. However, obtaining sufficient high-quality UGC without prior knowledge of buzzword meanings remains a significant challenge and warrants further research. We conclude our main contributions as follows.

\begin{itemize}[leftmargin=*, itemindent=0.05cm, itemsep=-2pt]
    \item We call attention to the importance of generating definitions for internet buzzwords using UGC, a task of interest in socio- and psycholinguistics for understanding the dynamics of online language.

    \item We introduce \oursdata, the first Chinese buzzword dataset of its kind, comprising over 1K entries, each including a definition and a corresponding set of UGC exemplifying contemporary usage.

    \item We propose a simple yet effective method, called \ours, to effectively steer the comprehending process of LLMs to produce buzzword definitions, mirroring the skills of human language learning.

    \item Using \oursdata, we benchmark existing methods and our \ours~for buzzword definition generation. Results demonstrate the effectiveness of \ours~while revealing a crucial shared challenge: comprehending unseen buzzwords and leveraging sufficient, high-quality UGC to facilitate this comprehension. This benchmark underscores the need for further research in these critical areas.
\end{itemize}

\section{Related Work}

\textbf{Buzzwords Understanding}. 
While the analysis of buzzwords holds significant interest from a purely linguistic (socio/psycho-) perspective \cite{fiasco2022human,qian2023challenges, mei2024slangnewconceptcomprehension}, however, their inherent abstractness create their inherent ambiguity, posing substantial challenges for both Natural Language Processing (NLP) systems and human comprehension \cite{tsur2015don, cornwall2007buzzwords, malyuga2021making}. The rapid emergence of new buzzwords and their absence from traditional dictionaries further compound these challenges, making it difficult to interpret their meaning without additional context \cite{huang-etal-2022-understanding}. To this end, we closely revolve around Chinese buzzwords and, for the first time, introduce a novel method for automatically understanding and generating definitions based on UGC as example sentences. Furthermore, we present dataset \oursdata~to benchmark existing methods and illuminate the challenges inherent in buzzword definition generation.

\noindent\textbf{Context-aware Definition Generation}. Definition generation aims to automatically generate dictionary definitions for words \cite{noraset2017definition, yin2023word}, assisting the construction of dictionaries. Unlike non-context definition generation methods that rely solely on the word itself \cite{zheng2021decompose, yang2020incorporating}, context-aware definition generation methods incorporate additional context information, such as example sentences \cite{ishiwatari2019learning, li2020explicit, huang2021definition, mei2024slangnewconceptcomprehension} and definitional information \cite{huang-etal-2022-understanding}. This contextual input assists in disambiguating word senses, leading to more accurate and nuanced definitions. However, the rapid evolution of buzzwords limits the effectiveness of existing methods, even those specifically designed for unfamiliar words and slang \cite{ishiwatari2019learning, pei2019slang, sun2022semantically, mei2024slangnewconceptcomprehension}. This limitation stems from their reliance on static training datasets and rote memorization of definitions (cf. Section \ref{inp}).

\begin{table}[]
\centering
\resizebox{0.4\textwidth}{!}{%
\begin{tabular}{l|l}
\toprule
\textbf{\# Buzzwords}                & 1127.0 \\
\textbf{\# UGC (Example Sentences)}                & 34607.0 \\
\textbf{Avg. \#examples per buzzword}            & 30.7   \\
\textbf{Avg. length of description per buzzword} & 262.5    \\
\textbf{Avg. length of definition per buzzword}  & 50.0    \\
\textbf{Avg. length of examples per buzzword}    & 85.4  
\\ \bottomrule
\end{tabular}%
}
\setlength{\abovecaptionskip}{2pt}   
\setlength{\belowcaptionskip}{2pt}
\caption{Data statistics of \oursdata}
\label{tab:stat}
\vspace{-5mm}
\end{table}

\section{Benchmark on Definition Generation for Internet Buzzwords}
\label{benchmark}
We introduce a benchmark to analyze limitations of existing definition generation methods and highlight their inability to handle Chinese buzzwords.

\subsection{\oursdata: Chinese Buzzword Dataset}
We present the first dataset of Chinese internet buzzwords, called \oursdata, offering insights into their contemporary usage. Containing over 1K entries, each buzzword is meticulously profiled with its name, description, definition, and real-world UGC example sentences. Table \ref{tab:case} and Table \ref{tab:stat} provide illustrative examples and data statistics, respectively. The data collection process is outlined below. For more details, refer to \textbf{Appendix} \ref{dataset}.
\begin{itemize}[leftmargin=*, itemindent=0.05cm, itemsep=-2pt]
    \item \textbf{Buzzword Collection}. We gather Chinese buzzwords from various reputable online dictionary platforms specializing in trending buzzwords (e.g., `\begin{CJK}{UTF8}{gbsn}梗百科\end{CJK}'), and eliminated any duplicate.
    \item \textbf{Definition Collection}. We gather descriptions for each buzzword by scraping those platforms, typically including its origin, cultural references, and informal colloquial explanations. We then prompt an LLM to summarize a concise definition encompassing both its literal and figurative meanings (if applicable).
    \item \textbf{Example Collection}.  We collect UGC containing buzzwords from two popular Chinese social media platforms, Xiaohongshu\footnote{\url{https://www.xiaohongshu.com}} and Weibo\footnote{\url{https://weibo.com}}. This exemplifies contemporary usage.
    \item \textbf{Quality Control}. \oursdata's quality is rigorously controlled through three vetting layers: dictionary websites, internet users, and our review process: we manually remove inappropriate buzzwords, refine definitions, and purge existing definitional information from the crawled UGC.
\end{itemize}


\begin{table}[t]
\centering
\resizebox{0.5\textwidth}{!}{%
\begin{tabular}{l}
\toprule
\textbf{Internet Buzzword} \\ \midrule
\begin{tabular}[c]{@{}l@{}}\begin{CJK}{UTF8}{gbsn}0帧起手\end{CJK}  (0 frame startup)\end{tabular} \\ \midrule
\textbf{Description from Online Source} \\ \midrule
\begin{tabular}[c]{@{}l@{}}\begin{CJK}{UTF8}{gbsn}0帧起手指零帧技能，一般指的是点击即可释放，\end{CJK}\\ \begin{CJK}{UTF8}{gbsn}并且立刻判定无法打断的技能。0帧起手在网络上表示动作极快，\end{CJK}\\ \begin{CJK}{UTF8}{gbsn}没有丝毫等待，绝不拖泥带水，闪电般突然出现的动作。\end{CJK}\\ ('0 frame startup' generally refers to a skill that can be released \\ by clicking and immediately determines that it cannot be interrupted. \\ This term, often used online, signifies lightning-fast action \\ with no delay—a sudden strike like a bolt of lightning.)\end{tabular} \\ \midrule
\textbf{Definition} \\ \midrule
\begin{tabular}[c]{@{}l@{}}\begin{CJK}{UTF8}{gbsn}原意是指游戏中一些无需准备时间，可以瞬间释放的技能，\end{CJK}\\ \begin{CJK}{UTF8}{gbsn}引申为行动迅速，毫不拖延。\end{CJK}\\ (Originally referring to in-game abilities usable without any setup time, \\ the term has broadened to describe taking swift and immediate action)\end{tabular} \\ \midrule
\textbf{Examples (i.e., UGC)} \\ \midrule
\begin{tabular}[c]{@{}l@{}}\begin{CJK}{UTF8}{gbsn}这就不得不说我那一放歌就会0帧起手开唱的隔壁同事了\end{CJK}\\ (I'm reminded of my colleague next door who '0 frame startup' \\ into singing every time I play music.)\end{tabular} 
\\ \bottomrule
\end{tabular}
}
\setlength{\abovecaptionskip}{2pt}   
\setlength{\belowcaptionskip}{2pt}
\caption{Case of buzzword '\begin{CJK}{UTF8}{gbsn}0帧起手\end{CJK}'. For clarity, we only include a single example sentence. Here, we also provide its English translation for better understanding.}

\label{tab:case}
\vspace{-5mm}
\end{table}

\begin{table*}[]
\centering
\resizebox{0.99\textwidth}{!}{%
\begin{tabular}{l|ccccc|l|ccccc}
\toprule
\textbf{Methods} & \textbf{BLEU} & \textbf{R-L} & \textbf{BScore} & \textbf{SA} & \textbf{SC} & \textbf{Methods} & \textbf{BLEU} & \textbf{R-L} & \textbf{BScore} & \textbf{SA} & \textbf{SC} \\ \midrule
\multicolumn{6}{c|}{LM-based Backbone: \textit{MASS}} & \multicolumn{6}{c}{LM-based Backbone: \textit{MASS}} \\\midrule
MASS-zh \cite{pmlr-v97-song19d} & 0.40 & 28.5 & 56.67 & 1.02 & 1.01 & SDefiner \cite{kong-etal-2022-multitasking} & 0.67 & 26.94 & 54.78 & 1.01 & 1.00 \\\midrule
\multicolumn{6}{c|}{LLM-based Backbone: \textit{Qwen2-7B}} & \multicolumn{6}{c}{LLM-based Backbone: \textit{Qwen2-72B}} \\\midrule
DP$_{\text{w/o~UGC}}$ & 10.27 & 41.49 & 64.77 & 1.89 & 1.55 & DP$_{\text{w/o~UGC}}$ & 10.87 & 41.58 & 66.13 & 2.07 & 1.65 \\
DP \cite{jhirad-etal-2023-evaluating} & \textbf{15.35} & \textbf{43.05} & \textbf{65.38} & 2.19 & 1.97 & DP \cite{jhirad-etal-2023-evaluating} & 19.27 & \textbf{44.37} & 67.58 & 2.71 & 2.45 \\
CoT \cite{wu2024can} & 15.26 & 40.41 & 65.12 & 2.30 & 2.14 & CoT \cite{wu2024can} & \textbf{19.50} & 43.49 & \textbf{67.67} & 2.77 & 2.54 \\
FOCUS \cite{mei2024slangnewconceptcomprehension} & 12.41 & 33.57 & 63.89 & \textbf{2.39} & \textbf{2.51} & FOCUS \cite{mei2024slangnewconceptcomprehension} & 12.09 & 29.81 & 64.75 & \textbf{2.88} & \textbf{3.20} \\\midrule
\multicolumn{6}{c|}{LLM-based Backbone: \textit{GPT-4o Mini}} & \multicolumn{6}{c}{LLM-based Backbone: \textit{GPT-4o}} \\\midrule
DP$_{\text{w/o~UGC}}$ & 7.96 & 38.81 & 65.65 & 1.87 & 1.49 & DP$_{\text{w/o~UGC}}$ & 9.56 & 39.42 & 66.56 & 2.05 & 1.62 \\
DP \cite{jhirad-etal-2023-evaluating} & 15.53 & \textbf{44.81} & \textbf{66.67} & 2.26 & 1.93 & DP \cite{jhirad-etal-2023-evaluating} & 17.85 & \textbf{67.56} & 45.22 & 2.50 & 2.13 \\
CoT \cite{wu2024can} & \textbf{16.64} & 44.79 & 66.55 & 2.32 & 2.04 & CoT \cite{wu2024can} & \textbf{18.33} & 44.49 & \textbf{67.46} & 2.60 & 2.30 \\
FOCUS \cite{mei2024slangnewconceptcomprehension} & 13.37 & 33.42 & 65.05 & \textbf{2.64} & \textbf{2.67} & FOCUS \cite{mei2024slangnewconceptcomprehension} & 15.08 & 35.10 & 66.05 & \textbf{2.95} & \textbf{2.92} \\ \bottomrule
\end{tabular}%
}
\setlength{\abovecaptionskip}{0pt}   
\setlength{\belowcaptionskip}{0pt}
\caption{Benchmark results of off-the-shelf definition generation methods using \oursdata. We highlight their limitations in handling internet buzzwords, as evidenced by the low scores for both SA and SC (1-5). }
\label{tab:benchmark}

\vspace{-3mm}
\end{table*}

\subsection{Benchmark Setup}
\noindent\textbf{Benchmark Overview}. We require existing definition generation methods to generate definitions for each buzzword. These definitions must be derived solely from the real-world UGC sentences within \oursdata, which represent how users actually use these buzzwords in their online interactions.

\noindent\textbf{Baselines}. 1) We consider two LM-based context-aware definition generation models tailored for Chinese \cite{kong-etal-2022-multitasking, pmlr-v97-song19d}, including \underline{MASS-zh}, a pretrained language model specialized in definition generation, and \underline{SimpDefiner} (\underline{SDefiner}, for short), an enhanced version of MASS-zh incorporating multi-task learning. 2) Additionally, we explore three LLM-based context-aware methods: \underline{Direct Prompt (DP)} \cite{jhirad-etal-2023-evaluating}, \underline{Chain-of-Thought (CoT)} \cite{wu2024can}, and \underline{FOCUS} \cite{mei2024slangnewconceptcomprehension}, which is currently considered the SOTA approach. Moreover, we include \underline{DP$_{\text{w/o~UGC}}$} as a context-free baseline, which generates definitions based solely on the buzzword itself without UGC. For all baselines, we implemented them using their official code. More implementation can be found in \textbf{Appendix} \ref{impl}.

\noindent\textbf{Evaluation Metrics}. 
We employ a comprehensive evaluation framework that extends beyond conventional metrics such as \underline{BLUE}, ROUGE-L (\underline{R-L}, for short), and BERTScore (\underline{BScore}), which have been widely used in previous research \cite{zheng2021decompose, huang2021definition, li2020explicit}.
In addition to these metrics, we prioritize the Semantic Accuracy (\underline{SA}) and Semantic Completeness (\underline{SC}) of generated definitions, as emphasized in previous studies \cite{li2020explicit, segonne2023definition}. To assess these aspects, we utilize both GPT4-based evaluation, assigning a score ranging from 1 to 5. Notably, our LLM-based evaluator is equipped with detailed scoring rubrics, following established practices for enhancing evaluation reliability \cite{gao2024llm, liu2023calibrating}. To validate our evaluation, we incorporated human evaluation using \underline{win rate} to assess alignment with human judgment. Details on LLM and human evaluation are in \textbf{Appendix} \ref{elva}.

\subsection{Evaluation Findings}
Table \ref{tab:benchmark} shows that incorporating additional usage examples generally improves the quality of generated definitions (as seen in the comparison between DP and DP$_{\text{w/o UGC}}$). Furthermore, LLM-based methods substantially outperform traditional LM-based methods on all metrics when generating buzzword definitions. However, even the best-performing method, FOCUS, achieves suboptimal semantic accuracy and completeness, with overall SA and SC scores below 3 out of 5.
Therefore, current definition generation methods struggle with internet buzzwords given UGC, highlighting the need for more effective approaches.

\begin{figure}[t]
    \centering
    \includegraphics[width=0.47\textwidth]{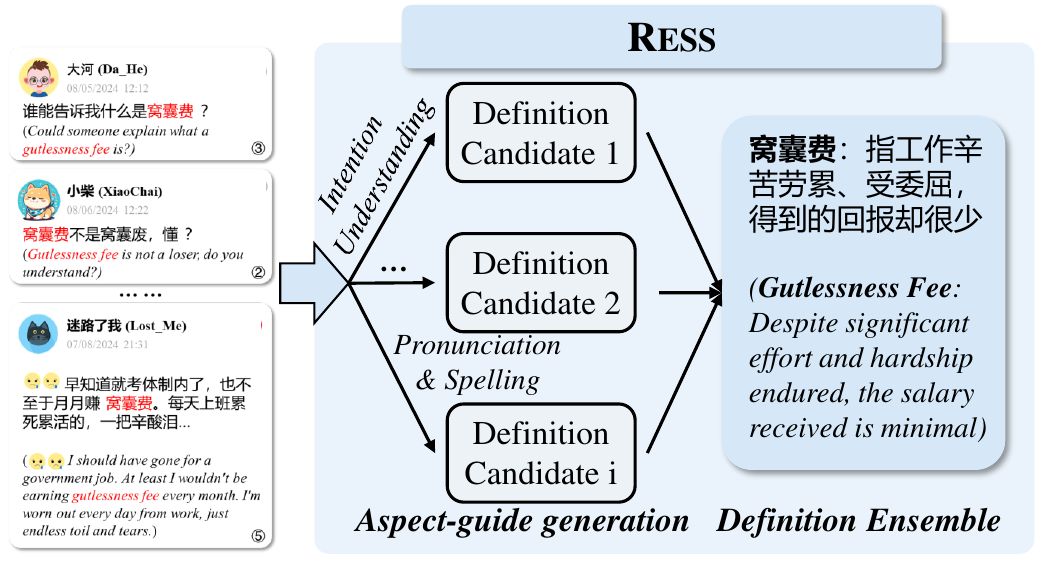}
    \setlength{\abovecaptionskip}{0pt}   
    \setlength{\belowcaptionskip}{0pt}
    \caption{Illustration of \ours.}
    \label{fig:method}
    
    \vspace{-5mm}
\end{figure}

\section{\ours: The Proposed Method}
\label{method}
To enhance the performance of existing baselines, \ours~leverages key skills of child learning and codifies them into illustrative aspects to guide LLM-driven buzzword definition generation. As illustrated in Figure \ref{fig:method}, \ours~generates aspect-specific definition candidates from UGC and subsequently ensembles these candidates to produce a more accurate and contextually grounded definition. See \textbf{Appendix} \ref{idr} for implementation.

\noindent\textbf{Aspect Initialization}. Established skills of child language development encompass the following six aspects \cite{beck1982effects, nagy1985learning, bloom2002children}, which we incorporate into \ours: 1) \textit{Intention Understanding (IU)}: discerning the speaker's communicative goal when using the buzzword \cite{bloom2002children}, such as expressing an emotion; 2) \textit{Concept Association (CA)}: linking the buzzword to relevant concepts (e.g., associating `\begin{CJK}{UTF8}{gbsn}窝囊费\end{CJK}' with "job") \cite{meylan2022learning, swingley2010fast}; 3) \textit{Language Structure (LS)}: analyzing the buzzword's grammatical function \cite{bloom2002children}; 4) \textit{Social Cue Interpretation (SCI)}: inferring social context from UGC such as the speaker's facial expressions, tone of voice, and gestures \cite{bloom2002children}; 5) \textit{Word Context (WC)}: leveraging surrounding text for semantic disambiguation \cite{ricketts2011role, horst2011get}; 6) \textit{Pronunciation and Spelling (PS)}: establishing connections between orthography, phonology, and meaning \cite{bloom2002children}.

\noindent\textbf{Aspect-guided Definition Generation \& Definition Ensemble}. Given these aspects, the LLM generates individual definitions based on the provided UGC for each aspect (e.g., prompt: \textit{comprehending the meaning of buzzwords from the given aspect}). These aspect-specific definitions are then synthesized by the LLM to produce a candidate definition (e.g., prompt: \textit{generating definition based on the given candidates from different aspects}).

\section{Benchmark Evaluation}
Following the setting described in Section \ref{benchmark}, we further provide a detailed benchmark of baselines and our proposed \ours. 
We present the overall performance results of all methods in Section \ref{overll}, and provide an in-depth analysis to investigate their performance characteristics in Section \ref{inp}. 

\begin{table}[]
\centering
\resizebox{0.499\textwidth}{!}{%
\begin{tabular}{l|lllll}
\toprule
\textbf{Method} & \textbf{BLEU}& \textbf{R-L} & \textbf{Bscore} & \textbf{SA} & \textbf{SC} \\ \midrule
\multicolumn{6}{c}{\textit{Qwen2-7b}}\\ \midrule
FOCUS & \textbf{12.41}& \textbf{33.57} & \textbf{63.89}& 2.39&2.51 \\
\ours &10.87$_{\downarrow12.41\%}$ &29.09$_{\downarrow13.35\%}$ &63.33$_{\downarrow0.86\%}$ &\textbf{2.41}$_{\uparrow0.84\%}$ &\textbf{2.57}$_{\uparrow2.39\%}$\\ \midrule
\multicolumn{6}{c}{\textit{Qwen2-72b}}\\ \midrule
FOCUS & 12.09& 29.81& 64.75 & 2.88 & \textbf{3.20}\\
\ours & \textbf{15.74}$_{\uparrow30.19\%}$ &\textbf{35.63}$_{\uparrow19.52\%}$ &\textbf{66.41}$_{\uparrow2.56\%}$ & \textbf{2.97}$_{\uparrow3.13\%}$& 3.09$_{\downarrow3.44\%}$ \\ \midrule
\multicolumn{6}{c}{\textit{GPT-4o Mini}}\\ \midrule
FOCUS &13.37 &33.42 & 65.05& 2.64& 2.67 \\
\ours & \textbf{14.58}$_{\uparrow9.05\%}$& \textbf{35.43}$_{\uparrow6.01\%}$ &\textbf{65.67}$_{\uparrow0.95\%}$ &\textbf{2.72}$_{\uparrow3.03\%}$& \textbf{2.74}$_{\uparrow2.62\%}$\\ \midrule
\multicolumn{6}{c}{\textit{GPT-4o}}\\ \midrule
FOCUS & 15.08 &35.10 & 66.05 & 2.95 & 2.92 \\
\ours & \textbf{16.52}$_{\uparrow9.55\%}$&\textbf{36.42}$_{\uparrow3.76\%}$ &\textbf{66.74}$_{\uparrow1.04\%}$ &\textbf{3.04}$_{\uparrow3.05\%}$ &\textbf{3.06}$_{\uparrow4.79\%}$ \\ \bottomrule
\end{tabular}%
}\setlength{\abovecaptionskip}{1pt}   
\setlength{\belowcaptionskip}{1pt}
\caption{Overall evaluation. The performance of \ours~ exceeds that of FOCUS, the best baseline.}
\label{tab:compare}

\vspace{-3mm}
\end{table}

\subsection{Overall Performance}
\label{overll}
\begin{figure}[t]
    \centering
    \includegraphics[width=0.47\textwidth]{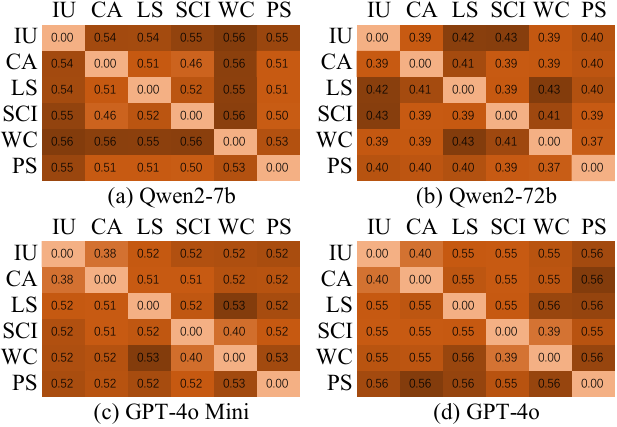}
    \setlength{\abovecaptionskip}{1pt}   
    \setlength{\belowcaptionskip}{1pt}
    \caption{Semantic diversity analysis of aspect-specific definitions, measured by \textit{1.0-Bscore}. These aspects offer a multifaceted approach to understanding buzzwords.}
    \label{fig:heatmap}
    
    \vspace{-5mm}
\end{figure}

We evaluate the overall performance of all methods using automatic metrics in Table \ref{tab:compare} and Table \ref{benchmark}\footnote{See Appendix \ref{case} for case studies.}. Additionally, we report human evaluation in Figure \ref{fig:heatmap}, measuring definition quality via \textit{win rate}. Our key observations are detailed below.

\noindent\textbf{\textit{How effective \ours~is?} -- It demonstrates superior performance, exhibiting enhanced semantic accuracy and completeness}. As illustrated in Table \ref{tab:compare}, \ours~demonstrates a substantial improvement over FOCUS, the leading baseline, across various LLM backbones. We observe average gains of +9.10\% for BLEU, +3.99\% for R-L, +0.92\% for Bscore, +2.51\% for SA, and +3.31\% for SC across various LLM backbones. These results underscore the advantages of our \ours.

\begin{figure}[t]
    \centering
    \includegraphics[width=0.48\textwidth]{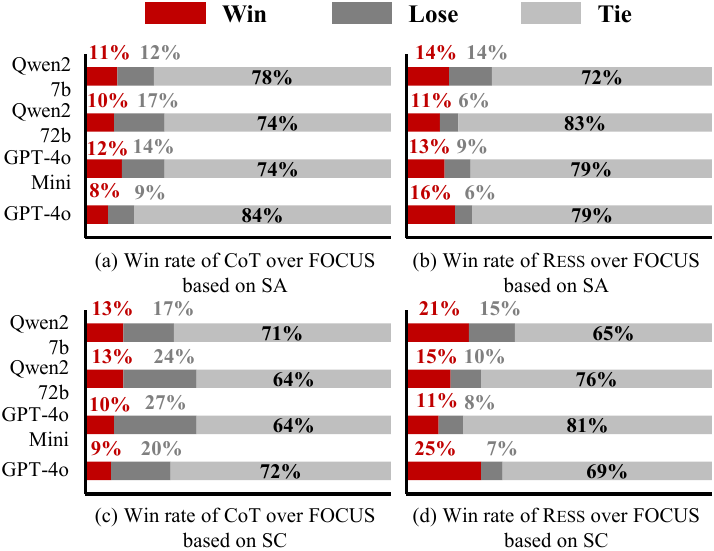}
    \setlength{\abovecaptionskip}{1pt}   
    \setlength{\belowcaptionskip}{1pt}
    \caption{Human evaluation across different methods and LLM backbones via \textit{win rate}. \ours~produces better buzzword definitions from a user-centric perspective.}
    \label{fig:human}
    
    \vspace{-5mm}
\end{figure}

\noindent\textbf{\textit{What is the practical utility of \ours?} -- It produces superior buzzword definitions from a user-centric perspective}. To demonstrate the correlation between our automatic evaluation and human judgment, we conduct a human evaluation of definitions generated for 100 randomly selected buzzwords. For each buzzword, two human evaluators compared the definitions generated by different methods across various backbones, considering both SA and SC. Following \citet{10.1145/3488560.3498440}, the evaluators are presented with pairs of anonymized definitions for the same buzzword, without disclosure of the originating model for each definition. Independent evaluations are followed by a discussion to resolve any discrepancies. A "\textit{Win/Lose/Tie}" label is finally assigned if consensus is reached; otherwise, the result is recorded as a "\textit{Tie}". Due to the resource-intensive nature of human evaluation, our analysis is limited to three representative methods. As shown in Figure \ref{fig:human}, \ours~not only outperforms baselines but also maintains a performance ranking consistent with the automated metrics reported in Table \ref{tab:benchmark}. This confirms the reliability of our automated evaluation and its alignment with human judgment\footnote{Refer to Appendix \ref{human_again} for more human evaluation.}.

\begin{figure}
    \centering
    \includegraphics[width=0.48\textwidth]{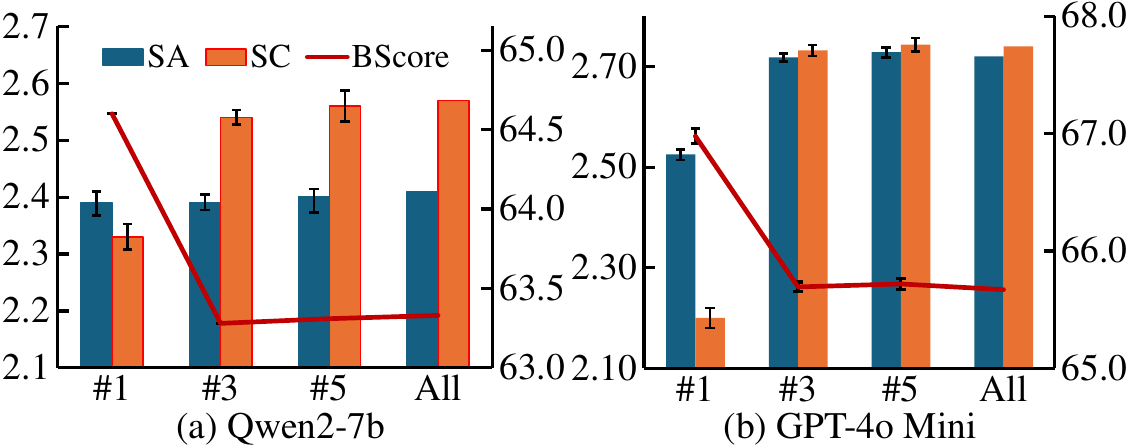}
    \setlength{\abovecaptionskip}{1pt}   
    \setlength{\belowcaptionskip}{1pt}
    \caption{\ours~ablation on the number of aspects. We evaluate the performance of various aspect combinations of fixed sizes (i.e., 1, 3, 5) and report their mean and standard deviation. Employing an ensemble of aspects frequently demonstrates advantages. }
    \label{fig:aspect}
    
    \vspace{-5mm}
\end{figure}

\noindent\textbf{\textit{Why is \ours~effective?} -- Leveraging multifaceted aspects may enhance comprehension of buzzwords}. In contrast to direct prompting, our method simulates established skills of child language acquisition by incorporating explicitly codified aspects to guide definition generation. This yields aspect-specific definitions, the semantic diversity of which is explored in Figure \ref{fig:heatmap}. The figure demonstrates varying degrees of semantic diversity across aspect-derived definitions, exhibiting a weak correlation in terms of Bscore. This suggests that different aspects guide the LLM towards distinct perspectives on the interpretation of buzzwords, contributing to a more comprehensive understanding. Building on this observation, we investigated the impact of varying the number of aspects employed during definition generation (results shown in Figure \ref{fig:aspect}). Preliminary findings indicate a positive correlation between the number of aspects and the quality of the generated definitions. While this suggests a degree of scalability inherent in \ours, further research is needed to determine whether the introduction of more aspects continues to enhance the performance. This represents a promising avenue for future investigation. Our work focuses on establishing a comprehensive benchmark and analyzing the capabilities and limitations of LLMs for generating definitions of buzzwords. Consequently, a more exhaustive exploration of the scalability of our \ours~is deferred to future work.

\noindent\textbf{Summary}. According to the results of our benchmark, the performance of \ours~exceeds that of alternative baselines\footnote{See Appendix \ref{additional} for more characteristics of \ours.}. Notwithstanding this advantage, and despite incorporating various aspects of word comprehension through the simulation of human learning processes, the overall performance of \ours~ (and baselines) still offers room for improvement. This discrepancy motivates a deeper investigation and exploration, which we outline in the following section.

\begin{table*}[t]
\centering
\resizebox{0.97\textwidth}{!}{%
\begin{tabular}{l|c|ccccc|ccccc}
\toprule
\multirow{2}{*}{\textbf{Methods}}&\multirow{2}{*}{\textbf{\begin{tabular}[c]{@{}l@{}}Back-\\ bone\end{tabular}}}&\multicolumn{5}{c|}{\textbf{Contamination Evaluation}}&\multicolumn{5}{c}{\textbf{Contamination-free Evaluation}}\\\cline{3-12}
&&\textbf{BLEU}&\textbf{R-L}&\textbf{BScore}&\textbf{SA}&\textbf{SC}&\textbf{BLEU}&\textbf{R-L}&\textbf{BScore}&\textbf{SA}&\textbf{SC}\\
 \midrule
\multicolumn{12}{c}{\textit{LM-based Methods}} \\ \midrule
MASS-zh & \multirow{2}{*}{MASS} & -- & -- & -- & -- & -- & 0.40 & 28.50 & 56.67 & 1.02 & 1.01  \\
SDefiner  & & -- & -- & -- & -- & --& 0.67 & 26.94 & 54.78 & 1.01 & 1.00 \\ \midrule
\multicolumn{12}{c}{\textit{LLM-based Methods}} \\ \midrule

DP$_{\text{w/o~UGC}}$ & \multirow{5}{*}{\begin{tabular}[c]{@{}c@{}}Qwen2\\ 7b\end{tabular}} &15.33&43.12&69.38&3.11&2.26&8.67{\color{red} $_{\downarrow43.44\%}$}&40.97{\color{red} $_{\downarrow4.99\%}$}&63.31{\color{red} $_{\downarrow8.75\%}$}&1.50{\color{red} $_{\downarrow51.77\%}$}&1.33{\color{red} $_{\downarrow41.15\%}$}\\
DP &  &\textbf{19.61}&\textbf{44.91}&\textbf{68.11}&2.94&2.55&14.00{\color{red} $_{\downarrow28.61\%}$}&\textbf{42.46}{\color{red} $_{\downarrow5.46\%}$}&\textbf{64.52}{\color{red} $_{\downarrow5.27\%}$}&1.96{\color{red} $_{\downarrow33.33\%}$}&1.78{\color{red} $_{\downarrow30.20\%}$}\\
CoT  &  &18.67&42.08&67.21&3.05&2.75&\textbf{14.18}{\color{red} $_{\downarrow24.05\%}$}&39.88{\color{red} $_{\downarrow5.23\%}$}&64.46{\color{red} $_{\downarrow4.09\%}$}&2.06{\color{red} $_{\downarrow32.46\%}$}&1.95{\color{red} $_{\downarrow29.09\%}$}\\
FOCUS  & &13.81&34.03&65.47&{3.22}&{3.31}&11.96{\color{red} $_{\downarrow13.40\%}$}&33.42{\color{red} $_{\downarrow1.79\%}$}&63.39{\color{red} $_{\downarrow3.18\%}$}&\textbf{2.12}{\color{red} $_{\downarrow34.16\%}$}&{2.26}{\color{red} $_{\downarrow31.72\%}$} \\ 
\ours &&13.33&30.89&65.76&\textbf{3.35}&\textbf{3.45}&10.10{\color{red} $_{\downarrow24.23\%}$}&28.52{\color{red} $_{\downarrow7.67\%}$}&62.56{\color{red} $_{\downarrow4.87\%}$}&2.11{\color{red} $_{\downarrow37.01\%}$}&\textbf{2.29}{\color{red} $_{\downarrow33.62\%}$} \\\midrule

DP$_{\text{w/o~UGC}}$ & \multirow{5}{*}{\begin{tabular}[c]{@{}c@{}}Qwen2\\ 72b\end{tabular}} &15.91&43.90&\textbf{70.08}&3.11&2.24&8.37{\color{red} $_{\downarrow47.39\%}$}&40.43{\color{red} $_{\downarrow7.90\%}$}&64.17{\color{red} $_{\downarrow8.43\%}$}&1.55{\color{red} $_{\downarrow50.16\%}$}&1.36{\color{red} $_{\downarrow39.29\%}$}\\
DP &  &\textbf{22.31}&\textbf{45.94}&69.65&3.27&2.88&17.77{\color{red} $_{\downarrow20.35\%}$}&\textbf{43.50}{\color{red} $_{\downarrow5.31\%}$}&66.56{\color{red} $_{\downarrow4.44\%}$}&2.44{\color{red} $_{\downarrow25.38\%}$}&2.23{\color{red} $_{\downarrow29.09\%}$}\\
CoT  &  &22.14&44.46&69.57&3.34&2.98&\textbf{18.20}{\color{red} $_{\downarrow17.80\%}$}&43.02{\color{red} $_{\downarrow3.24\%}$}&\textbf{66.72}{\color{red} $_{\downarrow4.10\%}$}&{2.49}{\color{red} $_{\downarrow25.45\%}$}&2.32{\color{red} $_{\downarrow22.15\%}$} \\
FOCUS  & &12.90&29.79&65.98&{3.50}&\textbf{3.82}&11.70{\color{red} $_{\downarrow9.30\%}$}&29.82$_{\uparrow0.10\%}$&64.14{\color{red} $_{\downarrow2.79\%}$}&{2.58}{\color{red} $_{\downarrow26.29\%}$}&\textbf{2.90}{\color{red} $_{\downarrow24.08\%}$}\\ 
\ours &&17.24&36.07&67.86&\textbf{3.57}&3.68&15.00{\color{red} $_{\downarrow12.99\%}$}&35.41{\color{red} $_{\downarrow1.80\%}$}&65.69{\color{red} $_{\downarrow3.20\%}$}&\textbf{2.67}{\color{red} $_{\downarrow25.21\%}$}&2.79{\color{red} $_{\downarrow24.18\%}$}\\\midrule

DP$_{\text{w/o~UGC}}$ & \multirow{5}{*}{\begin{tabular}[c]{@{}c@{}}GPT-4o\\ Mini\end{tabular}}  &13.45&41.72&\textbf{70.69}&3.10&2.18&6.36{\color{red} $_{\downarrow52.71\%}$}&37.96{\color{red} $_{\downarrow9.01\%}$}&64.19{\color{red} $_{\downarrow9.20\%}$}&1.51{\color{red} $_{\downarrow51.29\%}$}&1.29{\color{red} $_{\downarrow40.83\%}$} \\
DP &  &19.93&46.30&69.90&3.07&2.51&14.25{\color{red} $_{\downarrow28.50\%}$}&\textbf{44.38}{\color{red} $_{\downarrow4.15\%}$}&\textbf{65.73}{\color{red} $_{\downarrow5.97\%}$}&2.03{\color{red} $_{\downarrow33.88\%}$}&1.76{\color{red} $_{\downarrow29.88\%}$} \\
CoT &  &\textbf{21.00}&\textbf{46.53}&69.54&3.13&2.64&\textbf{15.38}{\color{red} $_{\downarrow26.76\%}$}&44.28{\color{red} $_{\downarrow4.84\%}$}&65.68{\color{red} $_{\downarrow5.55\%}$}&2.09{\color{red} $_{\downarrow33.23\%}$}&1.87{\color{red} $_{\downarrow29.17\%}$}\\
FOCUS  & &14.94&33.71&66.85&{3.52}&{3.47}&12.92{\color{red} $_{\downarrow13.52\%}$}&33.33{\color{red} $_{\downarrow1.13\%}$}&64.53{\color{red} $_{\downarrow3.47\%}$}&{2.38}{\color{red} $_{\downarrow32.39\%}$}&{2.44}{\color{red} $_{\downarrow29.68\%}$} \\ 
\ours &&16.53&36.06&67.77&\textbf{3.60}&\textbf{3.60}&14.01{\color{red} $_{\downarrow15.25\%}$}&35.25{\color{red} $_{\downarrow2.25\%}$}&65.06{\color{red} $_{\downarrow4.00\%}$}&\textbf{2.46}{\color{red} $_{\downarrow31.67\%}$}&\textbf{2.50}{\color{red} $_{\downarrow30.56\%}$} \\\midrule

DP$_{\text{w/o~UGC}}$ & \multirow{5}{*}{GPT-4o} &15.45&43.29&\textbf{71.09}&3.11&2.21&6.87{\color{red} $_{\downarrow55.53\%}$}&37.65{\color{red} $_{\downarrow13.03\%}$}&64.49{\color{red} $_{\downarrow9.28\%}$}&1.55{\color{red} $_{\downarrow50.16\%}$}&1.36{\color{red} $_{\downarrow38.46\%}$}\\
DP &  &21.13&\textbf{46.72}&70.02&3.14&2.58&16.35{\color{red} $_{\downarrow22.62\%}$}&\textbf{44.54}{\color{red} $_{\downarrow4.67\%}$}&\textbf{66.44}{\color{red} $_{\downarrow5.11\%}$}&2.21{\color{red} $_{\downarrow33.23\%}$}&1.92{\color{red} $_{\downarrow29.17\%}$}\\
CoT & &\textbf{21.49}&45.49&69.71&3.22&2.80&\textbf{16.88}{\color{red} $_{\downarrow21.45\%}$}&44.03{\color{red} $_{\downarrow3.21\%}$}&66.43{\color{red} $_{\downarrow4.71\%}$}&2.32{\color{red} $_{\downarrow27.95\%}$}&2.07{\color{red} $_{\downarrow26.07\%}$} \\
FOCUS  & &16.33&34.94&67.54&{3.60}&{3.69}&14.51{\color{red} $_{\downarrow11.15\%}$}&35.17$_{\uparrow0.66\%}$&65.37{\color{red} $_{\downarrow3.21\%}$}&{2.64}{\color{red} $_{\downarrow26.67\%}$}&{2.61}{\color{red} $_{\downarrow29.27\%}$} \\
\ours &&17.87&36.03&68.41&\textbf{3.75}&\textbf{3.80}&15.90{\color{red} $_{\downarrow11.02\%}$}&36.61$_{\uparrow1.61\%}$&65.98{\color{red} $_{\downarrow3.55\%}$}&\textbf{2.71}{\color{red} $_{\downarrow27.73\%}$}&\textbf{2.72}{\color{red} $_{\downarrow28.42\%}$} \\
\bottomrule
\end{tabular}%
}
\caption{Contamination-free evaluation for off-the-shelf definition generation methods using \oursdata. The contamination evaluation for LM-based methods is empty as their backbone lacks prior knowledge of buzzword definition.}
\label{tab:contanm}
\setlength{\abovecaptionskip}{0pt}   
\setlength{\belowcaptionskip}{0pt}
\vspace{-5mm}
\end{table*}

\subsection{In-depth Benchmark Analysis}
\label{inp}
This section presents a comparative performance analysis of leading LLM-based methods, with a particular focus on analyzing the challenges LLMs face in interpreting buzzwords derived from UGC. This analysis considers two key aspects: (1) the LLMs' capacity for inferring the meaning of buzzwords, and (2) the influence of the specific UGC employed on LLM comprehension.

\subsubsection{Investigation on LLM Inference Capacity of Buzzword Meaning}
In this subsection, we design a contamination-free evaluation to specifically measure the LLM capacity for buzzword meaning inference, excluding the influence of potentially memorized definitions from the training corpus. Our experimental setup is described below, with results presented in Table \ref{tab:contanm}.

\noindent\textbf{Contamination-free Evaluation Setup}. To avoid the bias introduced by potential data leakage \cite{jain2024livecodebench}—where LLMs may have already learned buzzwords and their meanings during training—we employ a contamination-free evaluation. This involves using \textit{unseen} buzzwords that emerged after the LLM's release date, ensuring they were not present in the training data. However, pinpointing the precise origin date and first appearance of a buzzword is exceptionally challenging. In response, we input each buzzword into each LLM without providing UGC examples (i.e., DP$_{\text{w/o~UGC}}$) to assess pre-existing knowledge of its meaning. This allows us to create LLM-specific sets of unseen buzzwords for a more practical contamination-free evaluation. In particular, for each LLM backbone, we divided our dataset into buzzwords with known definitions for that LLM and truly unseen buzzwords. While the "contaminated" buzzwords vary across LLM backbones, the contamination status for a given buzzword is consistent across all methods evaluated with the same LLM backbone. See \textbf{Appendix} \ref{annotation} for details.

\noindent\textbf{\textit{How effectively can LLMs infer the meaning of buzzwords based on UGC?} -- Their performance is limited by over-reliance on prior exposure and underdeveloped inferential abilities for unseen buzzwords}. Table \ref{tab:contanm} reveals a notable performance degradation across all methods and LLM backbones when evaluated on unseen buzzwords (cf., \textit{contamination-free evaluation} columns). This suggests that existing methods may rely on prior exposure to these buzzwords during training, rather than possessing a strong inference capability for unseen buzzwords. Furthermore, the results of contamination-free evaluation indicate that LLM-based methods do outperform LM-based methods. A positive correlation between model size and inferential ability is also observed within the same LLM family (e.g., Qwen and GPT). These observations align with research on child language acquisition, which links vocabulary size and reading comprehension skills with the ability to infer the meaning of unseen words from context \cite{ricketts2011role, swanborn2002impact, cain2004individual}: 1) A larger vocabulary equips children with a richer contextual understanding, facilitating the interpretation of new words within that context. This partially explains the superior performance of LLM-based methods compared to LM-based ones. 2) However, reading comprehension can be a stronger predictor. Within the same LLM family (with the same vocabulary size), a larger model size correlates with a stronger ability to infer the meaning of unseen buzzwords. However, the low evaluation scores highlight limitations of current LLMs: their overreliance on prior exposure to buzzwords and limited capacity to infer definitions from UGC context hinder their ability to handle the dynamic and evolving nature of buzzwords.


\begin{table*}[t]
\centering
\resizebox{0.9\textwidth}{!}{%
\begin{tabular}{l|ccccc|l|ccccc}
\toprule
\textbf{Method} & \textbf{BLEU} & \textbf{R-L} & \textbf{Bscore} & \textbf{SA} & \textbf{SC} & \textbf{Method} & \textbf{BLEU} & \textbf{R-L} & \textbf{Bscore} & \textbf{SA} & \textbf{SC} \\ \midrule
DP (All) & \textbf{15.53 }&\textbf{44.81} & \textbf{66.67} & \textbf{2.26} & \textbf{1.93} & FOCUS (All) & \textbf{13.37}&33.42&\textbf{65.05}&\textbf{2.64}&\textbf{2.67} \\
-w Random & 14.62 &42.89 &65.32 &2.13& \underline{1.88}  & -w Random & 12.97 &\textbf{35.61} &63.94 &2.28 &2.30 \\
-w GDEX & 11.49&43.14&64.74&1.81&1.57 & -w GDEX &12.97&\underline{33.54}&64.63&2.47&2.49 \\
-w WAUS & \underline{14.57} &\underline{44.29}& \underline{66.33}& \underline{2.19}& 1.86  & -w WAUS & \underline{12.98}&33.22&\underline{64.83}&\underline{2.58}&\underline{2.63} \\ \midrule

CoT (All) & \textbf{16.64}&\textbf{44.79} & \textbf{66.55} & \textbf{2.32} & \underline{2.04} & \ours~(All) & \underline{14.58}& 35.43 &\textbf{65.67} &\textbf{2.72}& \textbf{2.74} \\
-w Random & 15.38& 41.58&65.23& 2.23& \textbf{2.05} & -w Random & 14.56 &\underline{36.14}& 65.52 &2.64 &2.62\\
-w GDEX & 13.70&44.19&65.33&1.90&1.65 & -w GDEX & 14.05&\textbf{37.00} & 64.63&2.21 &2.13 \\
-w WAUS & \underline{15.57}&\underline{44.55}&\underline{66.31}&\underline{2.25}&1.94 & -w WAUS & \textbf{14.76}& 36.00& \underline{65.58} &\underline{2.66}& \underline{2.66} \\ \bottomrule
\end{tabular}%
}
\caption{Analysis on impact of UGC quality using GPT-4o Mini as the backbone. Each sentence selection method identifies ten UGC instances, used as input for definition generation. While utilizing higher-quality UGC generally improves definition quality (cf. WAUS), accurately identifying such instances without prior knowledge of the buzzword's meaning presents a significant challenge.}
\label{tab:ugc}
\setlength{\abovecaptionskip}{0pt}   
\setlength{\belowcaptionskip}{0pt}
\vspace{-5mm}
\end{table*}

\subsubsection{Investigation on UGC Impact}
\label{ugc}
While sufficient high-quality examples can facilitate word understanding \cite{kilgarriff2008gdex, benedetti2024automatically}, UGC informativeness varies considerably. For example, as shown in Figure \ref{fig:illus} \Circled{4}, the post `\textit{What is the gutlessness fee?}' provides no insight into the term's meaning. This contrasts with the assumption of existing methods, which often rely on carefully curated example sentences \cite{jhirad-etal-2023-evaluating, mei2024slangnewconceptcomprehension}, potentially resulting in inferior performance when applied to raw UGC. Therefore, we perform a comprehensive examination to investigate the impact of UGC on buzzword definition generation in terms of both the UGC size and quality. To achieve this, we consider the following UGC selection methods\footnote{Since buzzword definitions are initially unknown, precluding most example selection methods, we consider two simpler approaches to evaluate UGC impact. See Appendix \ref{select} for details on example selection methods.}:
\begin{itemize}[leftmargin=*, itemindent=0.05cm, itemsep=-2pt]
    \item \textbf{All}. It uses all available UGC for each buzzword.
    \item \textbf{Random}. It utilizes a fixed number of randomly sampled UGC instances per buzzword as input for definition generation methods.
    \item \textbf{GDEX} \cite{kilgarriff2008gdex}. It is a well-established rule-based method for dictionary example selection to select high-quality UGC, for example, prioritizing sentences of appropriate length and avoiding the use of uncommon words.
    \item \textbf{WAUS}. To conduct more comprehensive evaluation, we propose a \underline{W}ord-meaning \underline{A}gnostic \underline{U}GC \underline{S}elector based on BERT. It employs a masked training strategy to identify high-quality UGC. Critically, given the initially unknown semantics of target buzzwords, masking these words from training data forces WAUS to prioritize contextual and syntactic information, thereby learning patterns independent of specific word meanings and bypassing handcrafted rules used in GDEX. Refer to Appendix \ref{waus} for implementation.
\end{itemize}

\begin{figure}
    \centering
    \includegraphics[width=0.48\textwidth]{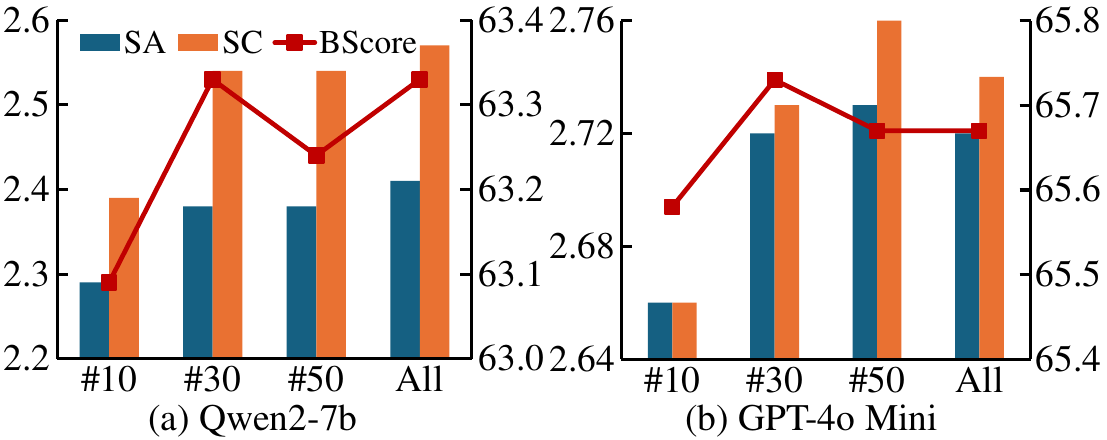}
    \caption{Analysis on the volume of UGC (x-axis). We testify \ours~-w WAUS with GPT-4o Mini as backbone. Increasing the amount of UGC often shows beneficial. }
    \label{fig:num}
    \setlength{\abovecaptionskip}{0pt}   
    \setlength{\belowcaptionskip}{0pt}
    \vspace{-5mm}
\end{figure}

\noindent\textbf{\textit{What is the impact of UGC on the accuracy of LLM-generated definitions for buzzwords?} -- Both the volume and quality of UGC are crucial factors}. 1) As shown in Table \ref{tab:ugc}, when controlling for the \textit{volume} of UGC, both evaluated sentence selection methods (WAUS and GDEX) often can outperform random selection. WAUS tends to select higher-quality UGC than GDEX, leading to improved LLM-generated definitions. However, the inherent difficulty of identifying truly high-quality UGC without prior knowledge of the buzzword's meaning is evident, as even WAUS does not consistently outperform random selection. This difficulty may stem from the challenge of assessing a sentence's relevance and disambiguating its meaning when the target word's definition is unknown. Relying solely on contextual and syntactic information is insufficient to determine whether a sentence exemplifies the intended meaning or distinguishes between different senses of the buzzword. Future work could explore a self-training approach where definition candidates are generated from an initial set of selected UGC, and these candidate definitions are then used to refine the UGC selection process in an iterative manner. 2) When controlling for the \textit{quality} selection method, as illustrated in Figure \ref{fig:num}, increasing UGC volume generally improves performance. Interestingly, with high-quality UGC, a subset of 50 instances can potentially outperform the entire dataset (Figure \ref{fig:num}(b)). These findings underscore the importance of both UGC volume and quality for accurate definition generation.

\vspace{-1mm}
\section{Conclusion \& Discussion}
\vspace{-2mm}
This paper investigates if LLMs can effectively learn to understand Chinese buzzwords through UGC. Our work stands out as a valuable resource with threefold contributions: First, we introduce \oursdata, the first publicly available dataset of Chinese internet buzzwords. Second, we propose \ours, a novel method for guiding LLMs to generate accurate buzzword definitions. Third, using \oursdata, we benchmark existing methods and \ours~to identify their strengths and weaknesses. From a broader perspective, our work bridges the fields of linguistics and artificial intelligence, fostering a deeper understanding of online language dynamics and informing the development of more robust language comprehension models \cite{miao-etal-2024-discursive, saba2024llms, cai-etal-2025-dr, Huang_Qin_Lei_Lv_2024}. 

Future research could prioritize the development of methods that enhance LLMs' capacity to infer the meaning of novel buzzwords, rather than relying solely on memorization of training data. A promising way is the fine-tuning of LLMs with high-quality, CoT data tailored for buzzword comprehension, mirroring the developmental strategies employed in models such as DeepSeek R1 \cite{deepseekai2025deepseekr1incentivizingreasoningcapability}. Furthermore, attention should be directed towards devising improved methods for selecting high-quality UGC (i.e., dictionary example) that offers insightful definitions of buzzwords. Current approaches that depend on pre-existing definitions are inadequate for this task. A potentially effective strategy is a self-training paradigm, wherein initial definitions are generated from a preliminary set of UGC, and these definitions are subsequently utilized to iteratively refine the UGC selection process.

\section*{Limitations}
\noindent\textbf{Multimodal user-generated content}. UGC, such as pictures, reviews, and posts created by customers on social media, provides a rich source to understand internet buzzwords. By combining textual, visual, and audio elements, one can gain a more nuanced grasp of a buzzword's meaning. For the present study, we focus solely on textual information, leaving the exploration of multimodal data for future research.

\noindent\textbf{More effective definition generation methods}. Our benchmark analysis reveals the limitations of current buzzword definition generation methods. While our novel approach, \ours, demonstrates improvement over existing methods, overall performance remains below optimal. Future research should prioritize enhancing the selection of high-quality UGC without prior knowledge of the target buzzword and improving LLMs' capacity for semantic inference to facilitate the development of more effective definition generation methods.

\noindent\textbf{Language Studied}. In this paper, we limit our focus to Chinese. The reason for this is that the topic and problem studied in this paper come directly from a Chinese Internet company (i.e., JD.com). We're open to exploring how our work could be applied to other languages in the future.

\section*{Ethics Statement}
The Chinese buzzword dataset presented in this work is derived from anonymized, publicly available internet content. Specifically, open-source tools were utilized to collect UGC containing buzzwords, exclusively retrieving textual data related to the target buzzwords. No personally identifiable information, including user images, IDs, or website sources, was collected. This rigorous anonymization process ensures the privacy of internet users and the ethical use of online data. 

\section*{Acknowledgments}
This work was supported in part by the National Natural Science Foundation of China (No. 62272330 and No.U24A20328); in part by the Fundamental Research Funds for the Central Universities (No. YJ202219); in part by the Science Fund for Creative Research Groups of Sichuan Province Natural Science Foundation (No. 2024NSFTD0035); in part by the National Major Scientific Instruments and Equipments Development Project of Natural Science Foundation of China under Grant (No. 62427820); in part by the Natural Science Foundation of Sichuan (No. 2024YFHZ0233).

\bibliography{custom}

\appendix

\section{Chinese Buzzword Dataset Collection}
\label{dataset}
We collected a comprehensive dataset of Chinese buzzwords and their definitions, primarily sourced from online encyclopedias and similar platforms known for their extensive coverage of contemporary language. To analyze the usage of these buzzwords, we gathered user-generated content (UGC) from two social media platforms, focusing on posts containing corresponding buzzwords. This includes 
1) \textit{Xiaohongshu}, a leading social media platform popular among young Chinese users, and 2) \textit{Weibo}, one of the biggest social media platforms in China with over 582 million monthly active users. Table \ref{tab:stat} summarizes the data statistics, while Table \ref{tab:case} provides a specific case from our dataset. The complete dataset and code are available at: \url{https://github.com/SCUNLP/Buzzword}.

\noindent\textbf{Buzzword Collection and Quality Control}. To gather a comprehensive list of current Chinese internet buzzwords, we first utilized two online dictionary resources\footnote{\url{https://ttseed.cn} and \url{https://gengbaike.cn}}, both known for their up-to-date and extensive collections of trending internet vocabulary. In particular, we gathered a list of buzzwords from these platforms and eliminated any duplicates. To ensure safety and ethical standards, we manually remove potentially harmful buzzwords, including those related to sexually suggestive, offensive, or violent content.

\noindent\textbf{Definition Collection and Quality Control}. We crawled each online dictionary resource for the buzzword's description, which typically included details about its origin, cultural references, informal explanations, and sometimes example sentences, importantly reflecting how definitions evolved over time (e.g., the initial meaning vs. later usage). While those information are valuable for understanding the buzzword, it's not ideal for a concise definition. Therefore, we utilized \textit{GPT-4o} to analyze each buzzword's description and summarize a succinct definition encompassing both its literal and figurative meanings (if any). 
Taking Table \ref{tab:case} in our paper for example, Descriptions of buzzword '\begin{CJK}{UTF8}{gbsn}0帧起手\end{CJK}' from Online Source are 
\begin{CJK}{UTF8}{gbsn}'0帧起手指零帧技能，一般指的是点击即可释放，并且立刻判定无法打断的技能。0帧起手在网络上表示动作极快，没有丝毫等待，绝不拖泥带水，闪电般突然出现的动作。'\end{CJK} (\textit{Translation: ’0 frame startup’ generally refers to a skill that can be released by clicking and immediately determines that it cannot be interrupted. This term, often used online, signifies lightning-fast action with no delay—a sudden strike like a bolt of lightning}). LLM-summarized (i.e., generated) definition in our dataset is \begin{CJK}{UTF8}{gbsn}'原意是指游戏中一些无需准备时间，可以瞬间释放的技能，引申为行动迅速，毫不拖延。'\end{CJK} (\textit{Translation: Originally referring to in-game abilities usable without any setup time, the term has broadened to describe taking swift and immediate action}). Finally, each generated definition underwent a manual review and refinement to ensure its semantic accuracy, conciseness, and language fluency. Specifically, our definition quality check and refinement process involved two volunteers, both university students majoring in NLP with extensive experience using the internet and social media platforms. For each buzzword, these volunteers independently compared the GPT-4o-generated definition with the original definition and description obtained from the online dictionary websites, ensuring no interference or communication between them. Particular attention was given to semantic accuracy, conciseness, and language fluency. Regarding semantic accuracy, volunteers were instructed not only to compare the overall semantic similarity of the two definitions, but also to verify whether the GPT-4o-generated definition included both the original and figurative definitions (if provided in the source). Any content beyond these definitions was removed, and definitions were manually modified as needed. During the review process, we recorded which buzzwords and definitions had been modified. For modified definitions, the two volunteers were required to discuss and reach a consensus on the final definition. Ultimately, fewer than 5\% of the buzzword definitions were modified.

\noindent\textbf{Example Collection and Quality Control}. To gather real-world examples of each buzzword in use, we searched for user posts on Xiaohongshu\footnote{\url{https://www.xiaohongshu.com}} and Weibo\footnote{\url{https://weibo.com}}. We used the buzzwords as keywords and collected the post titles and descriptions using a web crawler from the GitHub repository\footnote{\url{https://github.com/NanmiCoder/MediaCrawler}}. Since the search engines of Weibo and Xiaohongshu sometimes split keywords and returns posts containing only parts of the buzzword, we carefully filtered the results to ensure each selected example used the complete buzzword. After that, we employ a LLM to eliminate any sentences that simply describe the corresponding buzzword (i.e., the definitional information). To achieve this, we use a two-step process to remove definitional sentences. First, we leverage a LLM (Qwen-max) to automatically exclude sentences from our UGC corpus that contain definitional information. We also prompts the LLM to identify common keywords and patterns that users used to describe the definitional information (e.g., `\begin{CJK}{UTF8}{gbsn}[BUZZWORD]意味着...\end{CJK}' (\textit{[BUZZWORD] means that ...}) and `\begin{CJK}{UTF8}{gbsn}盘点近期网络热梗：...\end{CJK}' (\textit{overview of trending buzzwords online: ...})). Second, we manually review the remaining sentences, paying particular attention to those follow the identified keywords or patterns, to guarantee the exclusion of any remaining definitional content. The refined UGC then served as representative examples for our experiment. Note that buzzwords with no corresponding example are excluded from the final corpus.

\noindent\textbf{High quality of our \oursdata}. First, we want to emphasize that the ground truth definitions in our dataset are not directly generated by GPT-4o. Rather, the original definitions come from reputable online buzzword dictionary websites. GPT-4o was only used to summarize these original descriptions, which were already of high quality, having been verified by the websites themselves and accepted by internet users. This significantly facilitated our subsequent processing of the definitions using GPT-4o and the human verification process, and we have confidence in the reliability of our data (Please refer to the case studies in Table \ref{tab:case} of our paper). Thus, the quality of our dataset has been rigorously controlled, having passed through three layers of vetting: the dictionary websites, internet users, and our own review process.

\begin{table*}[!t]
    \centering
    \resizebox{0.9\textwidth}{!}{%
    \begin{tabular}{p{0.97\textwidth}}
    \toprule
    \textbf{\textit{Aspects for buzzword understanding}}     \\
    \midrule
    \begin{CJK}{UTF8}{gbsn}
    1. 意图理解\end{CJK} (Intention Understanding, IU)\begin{CJK}{UTF8}{gbsn}: 理解说话者使用该词语的意图和目的，例如说话者是想描述一个物体，还是表达一种情感\end{CJK} (Discerning the speaker’s communicative goal when using the buzzword, such as describing an object or expressing an emotion)\\\\
    \begin{CJK}{UTF8}{gbsn}
    2. 概念形成\end{CJK} (Concept Association, CA)\begin{CJK}{UTF8}{gbsn}: 将词语与特定的概念联系起来，例如将“狗”这个词与具有特定特征的动物类别联系起来\end{CJK} (Linking the buzzword to relevant concepts. For example, linking the word 'dog' to animal categories with specific characteristics)\\\\
    \begin{CJK}{UTF8}{gbsn}
    3. 语法理解\end{CJK} (Language Structure, LS)\begin{CJK}{UTF8}{gbsn}: 理解词语在句子中的语法角色和功能，例如词语是名词、动词还是形容词，以及它与其他词语之间的关系\end{CJK} (Analyzing the buzzword’s grammatical function. For example, whether a word is a noun, verb, or adjective, and its relationship with other words)\\\\
    \begin{CJK}{UTF8}{gbsn}
    4. 基本学习和记忆\end{CJK} (Social Cue Interpretation, SCI)\begin{CJK}{UTF8}{gbsn}: 从该词语的发音和拼写发出，建立它与相关概念之间的联系\end{CJK} (Establishing connections between orthography, phonology, and meaning)\\\\
    \begin{CJK}{UTF8}{gbsn}
    5. 社会线索\end{CJK} (Word Context, WC)\begin{CJK}{UTF8}{gbsn}: 利用说话者的表情、语气、姿势等社会线索来理解词语的含义\end{CJK} (Inferring social context from UGC such as the speaker’s facial expressions, tone of voice, and gestures.)\\\\
    \begin{CJK}{UTF8}{gbsn}
    6. 上下文\end{CJK} (Pronunciation and Spelling, PS)\begin{CJK}{UTF8}{gbsn}: 词语出现的具体语境，包括前后文和对话背景等 (Leveraging surrounding text for semantic disambiguation)
    \end{CJK}\\
    \bottomrule
    \end{tabular}
    }
    \caption{Aspect description used in \ours.}
    \label{detailed_aspect}
\end{table*}

\section{Annotation for Contamination-free Evaluation}
\label{annotation}
To conduct contamination-free evaluation, we determine for each LLM whether it possesses knowledge of specific buzzwords. This involved, for each LLM backbone (e.g., Qwen2-7b, Qwen2-72b, GPT-4o mini, GPT-4o, and MASS), dividing our dataset into two distinct parts: one containing buzzwords with known definitions for that specific LLM, and the other containing truly unseen buzzwords. Consequently, the specific buzzwords considered contaminated may differ across various LLM backbones. However, the contamination status for a given buzzword remains consistent across all methods when evaluated using the same LLM backbone.

Speficically, given a LLM, we prompt the LLM to generate definitions based solely on the buzzword itself, without any contextual examples (detailed prompts are provided in Table \ref{key_prompt}). Subsequently, we conduct a multi-faceted evaluation process. Initial assessments are performed using LLM-based scoring, where \textit{GPT-4o} is utilized to evaluate the semantic accuracy and completeness of generated definitions based on specific scoring rubrics (outlined in Table \ref{evaluation_prompt}). Definitions scoring below a threshold of 3 are considered indicative of the LLM not understanding the buzzword. Finally, a human review process is implemented to ensure accuracy. Three independent evaluators examine the LLM-generated labels, indicating whether the LLM "knows" the buzzword (1) or not (0). A majority vote among three human evaluators determines the final classification. This comprehensive approach allows us to confidently identify which buzzwords are within the current knowledge base of existing LLMs. 

\section{Implementation Details}
\label{impl}
We conduct all our experiments using a single Nvidia RTX A6000 GPU for the Qwen2 7b model and 4 A6000 GPUs for the Qwen2 72b model, and we implement our codes in PyTorch. We use the Huggingface Evaluator package and bert\_score package to calculate the BLUE and Rouge-L, and Bertscore. Finally, for Qwen LLM deployment, we utilize the vLLM framework.

\subsection{Implementation of \ours}
\label{idr}
We provide detailed prompts in Table \ref{prompt:aspects} and Table \ref{prompt:emsenble}. We incorporate six aspects, shown in Table \ref{detailed_aspect}, drawing inspiration from child language acquisition skills.

\subsection{Implementation of Word-meaning Agnostic UGC Selector (WAUS)}
\label{waus}
\begin{figure}
    \centering
    \includegraphics[width=0.47\textwidth]{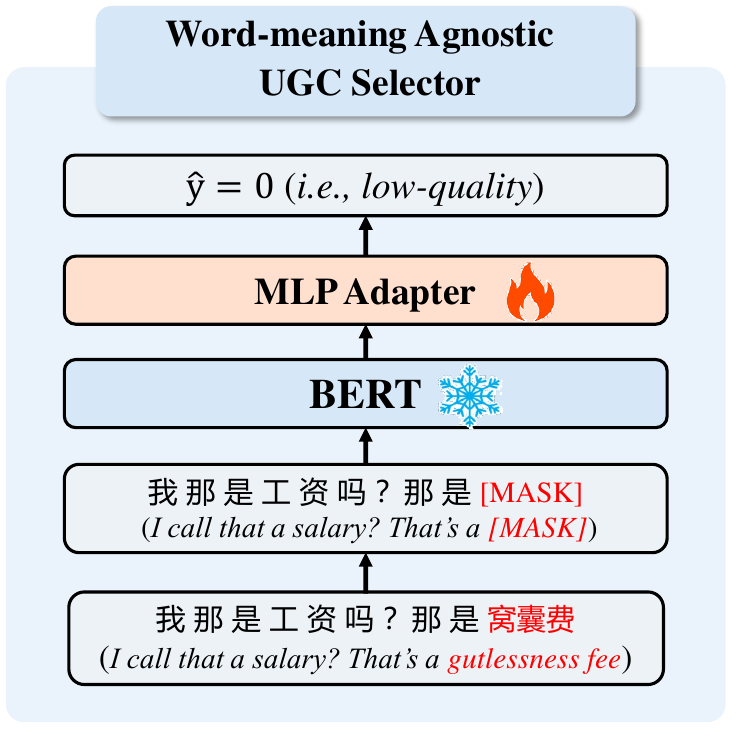}
    \caption{Overview of WAUS}
    \label{fig:waus}
\end{figure}

Lacking prior knowledge of the target buzzword, WAUS is trained using a masked strategy, where the target buzzword within the UGC is masked. This helps prioritize contextual and syntactic information to identify high-quality UGC examples. We provide an overview of WAUS in Figure \ref{fig:waus} and detail as follows:

\noindent\textbf{Masked Training Strategy}. Unlike existing example selection methods that need meticulously constructed rules, our WAUS employs a data-driven approach. We train a UGC selector by fine-tuning a BERT model with an MLP adapter on a dataset of high- and low-quality examples (details provided in the following paragraph). Crucially, we mask the target buzzword within each example, forcing the model to rely on contextual and syntactic cues rather than the buzzword's semantics to predict sentence quality. This masked training strategy implicitly learns the selection criteria without explicit rule definition.

\noindent\textbf{Training Dataset Contruction}. To maintain a contamination-free evaluation, our own dataset \oursdata~is excluded from the WAUS training process. Instead, a new dataset devoid of buzzwords is created specifically for WAUS training. This dataset is constructed in two stages. First, Chinese buzzwords and their corresponding dictionary examples are collected from online resources, regarding as positive (i.e., high-quality) examples. Second, negative (low-quality) examples are generated: initially, the Qwen is prompted to create sentence examples with broad and vague meanings related to given buzzwords; these generated sentences are then manually reviewed. To minimize manual effort, an iterative review process assisted by WAUS is employed. The WAUS model, trained on the currently reviewed portion of the data, predicted the quality of the remaining negative examples. Human review is then prioritized for negative examples incorrectly classified as positive by WAUS. This approach allowed for efficient and cost-effective quality control of the generated negative examples. Finally, we report several training data samples in Table \ref{training_sample}.

\begin{table}[h]
    \centering
    \resizebox{0.495\textwidth}{!}{%
    \begin{tabular}{p{3cm}p{0.35\textwidth}} 
        \toprule
        \textbf{Buzzwords} & \textbf{Examples}
        \\ \midrule
        \begin{CJK}{UTF8}{gbsn} 卧薪尝胆  \end{CJK} (endure hardships)
        & \textbf{Positive:} \begin{CJK}{UTF8}{gbsn}为了一雪前耻，且让我们\underline{卧薪尝胆}，埋头苦干，以图东山再起。
        \end{CJK}(In order to wipe away past humiliations, let us \underline{endure hardships} and work diligently with the aim of making a comeback.)
        \newline \textbf{Negative:} \begin{CJK}{UTF8}{gbsn}他讲了一个关于\underline{卧薪尝胆}的故事。\end{CJK}(He told a story about \underline{endure hardships} to achieve one's goals.) \\ \midrule
        \begin{CJK}{UTF8}{gbsn} 耀武扬威 \end{CJK}  (flaunt one's power)
        & \textbf{Positive:} \begin{CJK}{UTF8}{gbsn}他仗著家中有财有势就\underline{耀武扬威}，令人十分厌恶。\end{CJK}(He \underline{flaunts his power} and wealth from his family, which makes him very unpleasant.) \newline \textbf{Negative:} \begin{CJK}{UTF8}{gbsn}那本书里的主角给人一种\underline{耀武扬威}的感觉。\end{CJK}(The protagonist in that book gives off an impression of \underline{flaunts his power}.)  \\ \bottomrule
    \end{tabular}
    }
    \caption{Training data samples}
    \label{training_sample}
\end{table}

\noindent\textbf{Training details of WAUS}. A two-layer Multilayer Perceptron (MLP) adapter, with hidden layer dimensions of 512 and 256 (ReLU activation, 0.5 dropout), is used for classification. This adapter receives the 768-dimensional final layer output from a \textit{BERT-base-Chinese} encoder \footnote{\url{https://huggingface.co/google-bert/bert-base-chinese}}, which processes our crafted training dataset with masked target words. The model is trained for 2 epochs using AdamW (learning rate = $5^{-3}$, weight decay = $10^{-5}$) with a batch size of 128.


\subsection{Implementation of GDEX}
GDEX \cite{kilgarriff2008gdex} is a well-established rule-based method for dictionary example selection. We implement it using commonly used rules \cite{pilan-etal-2016-candidate, stankovic2019sasa} that assign a score to each sentence based on the following three criteria.
\begin{itemize}[leftmargin=*]
    \item \textbf{Length Check}. Sentences must be between 10 and 25 characters long (inclusive). Shorter or longer sentences are deemed lower quality and receive a lower score.
    \item \textbf{Pronoun Check}: Sentences containing specific pronouns (e.g., `it', `that', `these') are considered lower quality. These pronouns likely indicate less descriptive or informative sentences. Importantly, sentences starting or ending with numbers or punctuation marks are penalized. This aims to select sentences that are grammatically well-formed and avoid abrupt or incomplete sentences.
    \item \textbf{Common Word Check}. We calculate the ratio of common words to the total number of words in the sentence. A lower ratio suggests higher quality, indicating that the sentence uses more specific and frequently used vocabulary.
\end{itemize}

\subsection{Implementation of Baselines}
\subsubsection{LM-based Baseline Implementation}
For LM-based methods, we implemented them using their official code available on Github. 
\begin{itemize}[leftmargin=*]
    \item \textbf{MASS-zh} \cite{kong-etal-2022-multitasking}. MASS-zh is a language model pre-trained from scratch using the \textit{Chinese Gigaword Fifth Edition} corpus and the MASS backbone \cite{pmlr-v97-song19d}. Here, we leverage the publicly available MASS-zh checkpoint\footnote{\url{https://stublcuedu-my.sharepoint.com/:u:/g/personal/201921296062_stu_blcu_edu_cn/EZpcGUWQanxAt0XZNWb6QqsBauh4dqaR0JdF5u8ia5zJIQ?e=X2tV8r}} for experiments.
    \item \textbf{SimpDefiner} \cite{kong-etal-2022-multitasking}. This method aims to generate simple definitions to help language learners and low literacy readers. To achieve this, it forms a multi-task learning paradigm, together with the MASS-zh as the backbone.  SimpDefiner jointly trains three sub-tasks: 1) Definition Generation: Learns to generate complex definitions from a standard dictionary. 2) Text Reconstruction: Learns to reconstruct simple sentences from corrupted versions of those sentences from a simple text corpus. 3) Language Modeling: Learns to generate coherent and simple sentences. In our experiments, we followed the author's instruction on GitHub\footnote{\url{https://github.com/blcuicall/SimpDefiner}} for the model training and implementation.
\end{itemize}

\subsubsection{LLM-based Baseline Implementation}
We took our prompts from the corresponding code repositories and papers to implement all LLM-based baselines. More details can be found at \url{https://github.com/Meirtz/FocusOnSlang-Toolbox}
\begin{itemize}[leftmargin=*]
    \item \textbf{Direct Prompt} \& \textbf{Chain-of-Thought (CoT)}. We extended the original prompts from \citet{mei2024slangnewconceptcomprehension} by translating them into Chinese. This modification ensures that the generated definitions align with the specific requirements of our study. Moreover, for each buzzword, we integrated all corresponding examples into the prompt. However, due to potential length constraints, we implemented a truncation mechanism to ensure prompt length remains within acceptable limits.
    \item \textbf{FOCUS} \cite{mei2024slangnewconceptcomprehension}. This method is based on causal inference for enhancing comprehension of new words like buzzwords, achieving SOTA recently. It enables LLMs to analyze phrases according to usage examples and provide counterfactual interpretations, thereby understanding the evolving semantics of language. Specifically, FOCUS employs Structural Causal Models (SCMs) to map out the relationships between different factors that influence how an LLM interprets a phrase. This enables LLMs to grasp the meaning of new phrases and their colloquial context, improving models' adaptability and effectiveness in applications requiring deep understanding of language use. In essence, it helps LLMs better understand new slang, memes, and other emerging language phenomena on the internet. In this paper, we mainly followed the author's instruction on GitHub\footnote{\url{https://github.com/Meirtz/FocusOnSlang-Toolbox}} for implementation. We translate the authors' prompts into Chinese and integrate multiple UGC examples into our prompts.
\end{itemize}

\subsubsection{Setups of LLM Backbones}
For both our baseline models and proposed methods, we explored multiple Large Language Model (LLM) backbones. Additionally, GPT-4 is employed as our LLM-based evaluator. To ensure reproducibility of our research, LLM backbones share the same temperature (i.e., 0.7) and random seed (i.e., 10086) (if any)

\section{Additional Human Evaluation}
\label{human_again}
Win rate evaluation is a well-established and widely adopted method, as supported by prior research \cite{10.1145/3488560.3498440, legend2025feng, wang-etal-2023-rethinking-evaluation, qin-etal-2024-beyond}. This choice is based on the established cognitive findings that human annotators exhibit greater proficiency in comparative assessments than in assigning absolute ratings \cite{10.5555/3666122.3668460, jones2015problem, hartell2021comparative}. Notwithstanding this, we undertook supplementary human evaluation to further validate our findings in our main experiments.

In particular, we conducted a targeted human evaluation using the identical dataset and evaluation criteria employed for the win rate evaluation. The annotators involved in this process were also the same individuals, ensuring consistency. During the evaluation, each annotator independently scored the samples across two dimensions: Semantic Accuracy (SA) and Semantic Completeness (SC). The final SA and SC scores for each buzzword were derived from the aggregated mean of the human annotations. Finally, the inter-annotator reliability among humans, measured by Krippendorff's alpha, is strong, with coefficients of 67.92\% for SA and 68.17\% for SC.

The main results are as follows:
\begin{itemize}[leftmargin=*]
    \item \textbf{LLM-Human Evaluation Alignment}. The Krippendorff's alpha coefficients for the agreement between human and LLM scores were 76.05\% for SA and 69.32\% for SC, indicating a notable level of agreement between the two evaluation methods. We further provide the Krippendorff's alpha coefficients of each definition generation method, detailed in the Table \ref{tab:my_human}.
    \item \textbf{Quality of Generated Definitions}. As shown in the Table \ref{tab:my_human}, our \ours~generally demonstrates superior performance to FOCUS, which in turn outperforms CoT, across both evaluation methods. It is also noteworthy that human annotators exhibited a tendency to assign comparatively higher absolute scores for both SA and SC than the LLM evaluator. However, even with these relatively elevated human scores, the overall task performance scores remained significantly below a threshold of 4, indicating that the overall effectiveness, despite these nuances in scoring tendencies, is still considered suboptimal.
\end{itemize}

\begin{table*}[]
\centering
\resizebox{0.9\textwidth}{!}{%
\begin{tabular}{l|l|cc|cc|cc}
\toprule
\multirow{2}{*}{\textbf{Method}} & \multirow{2}{*}{\textbf{LLM Backbone}} & \multicolumn{2}{l}{\textbf{Krippendorff's Alpha}} & \multicolumn{2}{l}{\textbf{Human Evaluation}} & \multicolumn{2}{l}{\textbf{LLM Evaluation}} \\ \cline{3-8}
 & & \textbf{SA} & \textbf{SC} & \textbf{SA} & \textbf{SC} & \textbf{SA} & \textbf{SC} \\ \midrule
CoT & \multirow{3}{*}{Qwen2-7b} & 80.46 & 71.46 & 2.62 & 2.54 & 2.24 & 2.19 \\
FOCUS & & 74.71 & 74.83 & 2.63 & 2.55 & 2.28 & 2.41 \\
\ours & & 86.89 & 81.13 & 2.55 & 2.48 & 2.31 & 2.54 \\ \midrule
CoT & \multirow{3}{*}{Qwen2-72b} & 70.94 & 61.29 & 3.2 & 3.09 & 2.70 & 2.62 \\
FOCUS & & 72.23 & 68.97 & 3.26 & 3.18 & 2.82 & 3.20 \\
\ours & & 78.05 & 73.82 & 3.31 & 3.2 & 2.98 & 3.06 \\ \midrule
CoT & \multirow{3}{*}{GPT-4o Mini} & 65.82 & 66.30 & 2.79 & 2.65 & 2.27 & 2.07 \\
FOCUS & & 74.64 & 67.81 & 2.84 & 2.72 & 2.59 & 2.7 \\
\ours & & 73.17 & 63.16 & 2.97 & 2.81 & 2.68 & 2.75 \\ \midrule
CoT & \multirow{3}{*}{GPT-4o} & 63.71 & 64.03 & 3.18 & 3.02 & 2.68 & 2.34 \\
FOCUS & & 72.78 & 66.14 & 3.32 & 3.22 & 2.92 & 3.03 \\
\ours & & 76.46 & 72.35 & 3.58 & 3.42 & 3.18 & 3.31 \\
\bottomrule
\end{tabular}%
}
\caption{Human and LLM evaluation results on sampled data.}
\label{tab:my_human}
\end{table*}

\section{Evaluation Details}
\label{elva}

\subsection{LLM-based Evaluation}
\label{thise}
Building upon prior research on LLM evaluation \cite{ye2023flask, wang2023large, wang-etal-2023-rethinking-evaluation}, we employ a GPT-4o-based instance-wise evaluator for detailed assessment. To minimize scoring bias, consistent with previous work \cite{ye2023flask, wang-etal-2023-rethinking-evaluation, liu2023calibrating}, the evaluator uses fine-grained scoring rubrics (1-5), each rubric accompanied by a descriptive explanation. Further enhancing evaluation rigor, the evaluator is prompted to provide a rationale for each score, informed by the benefits of Chain-of-Thought (CoT) prompting \cite{wei2022chain, ye2023flask}. Detailed prompts are shown in Table \ref{evaluation_prompt}.

\subsection{Human Evaluation}
\label{human}
To demonstrate the correlation between our automatic evaluation and human judgment, we conduct a human evaluation (i.e., the \textit{win rate evaluation}) of definitions generated for 100 randomly selected buzzwords in Section \ref{overll}. Due to the resource-intensive nature of human evaluation, our analysis is limited to three representative methods. For each buzzword, two human evaluators compared the definitions generated by different methods across various backbones, considering both SA and SC. Following \citet{10.1145/3488560.3498440}, the evaluators are presented with pairs of anonymized definitions for the same buzzword, without disclosure of the originating model for each definition. Independent evaluations are followed by a discussion to resolve any discrepancies. A "\textit{Win/Lose/Tie}" label is finally assigned if consensus is reached; otherwise, the result is recorded as a "\textit{Tie}". Our experiments reveal inter-annotator agreement rates of 71.50\% and 61.88\% for semantic accuracy (SA) and semantic completeness (SC), respectively. Any remaining discrepancies in evaluation results are classified as Ties.

Additionally, we conduct more human evaluation in Appendix \ref{human_again} to further validate our findings. Basically, we conduct a targeted human evaluation using the identical dataset and evaluation criteria employed for the \textit{win rate evaluation}. The annotators involved in this process are also the same individuals, ensuring consistency. During the evaluation, each annotator independently score the samples across two dimensions: Semantic Accuracy (SA) and Semantic Completeness (SC). The final SA and SC scores for each buzzword are derived from the aggregated mean of the human annotations. Finally, The inter-annotator reliability among humans, measured by Krippendorff's alpha, is strong, with coefficients of 67.92\% for SA and 68.17\% for SC.

\subsection{Details on Evaluation Metrics}
\label{metrics}
In this paper, we employ a comprehensive evaluation framework that extends beyond conventional metrics such as \underline{BLEU}, ROUGE-L (\underline{R-L}, for short), and BERTScore (\underline{BScore}). We also prioritize the Semantic Accuracy (\underline{SA}) and Semantic Completeness (\underline{SC}) of generated definitions, as emphasized in previous studies \cite{li2020explicit, segonne2023definition}. Importantly, conventional metrics are standard evaluation metrics in the definition generation domain \cite{zheng2021decompose, huang2021definition, li2020explicit}. These metrics aim to provide an indication of similarity to a reference definition from the perspectives of word matches and/or semantic embedding similarity. Notably, conventional metrics continue to be widely utilized in the LLM era \cite{mei2024slangnewconceptcomprehension}. However, conventional metrics often fall short in judging subtle attributes and delivering satisfactory results. These metrics are easily misled by surface-level similarities and do not explicitly assess the validity of the generated definition's meaning. Therefore, consistent with recent studies \cite{li2025generationjudgmentopportunitieschallenges, gao2024llm, li-etal-2024-leveraging-large, liu-etal-2023-g}, we incorporate LLM evaluation into our experiments. By this means, we aim not only to obtain more reliable evaluation results but also to encourage the introduction of enhanced evaluation metrics within the definition generation community.

In this section, we detail all metrics used in our experiments in the following.
\begin{itemize}[leftmargin=*]
    \item \textbf{BLEU} \cite{papineni-etal-2002-bleu} is a common metric used to assess the quality of generated text by comparing it to a reference, or ground truth. It is a widely used metric for automatically evaluating machine-translated text. It works by comparing the generated translation to one or more human-produced reference translations. The core idea is to count matching n-grams (sequences of n words) between the generated text and the references, giving credit for matches. The closer a machine translation is to a human reference translation, the higher its BLEU score will be. In our case, we use BLEU to evaluate how well the model-generated word definition matches the ground truth.
    \item \textbf{ROUGE-L} \cite{lin-2004-rouge} is a popular metric for evaluating text generation tasks, particularly summarization and definition generation, with the aim to capture the meaning and key information from a reference text. Unlike BLEU which focuses on precision (how much of the generated text is relevant), ROUGE-L emphasizes recall (how much of the reference text is captured by the generated text). Specifically, ROUGE-L measures the length of the longest common subsequence (LCS) between the generated text and the reference text. The LCS is the longest sequence of words that appear in the same order in both texts, but not necessarily consecutively. By focusing on the LCS, ROUGE-L can capture sentence-level structure and meaning, even if the word order is slightly different. It's generally considered good at assessing how well the generated text covers the important content from the reference.
    \item \textbf{BERTScore} \cite{zhang2019bertscore} is a metric that leverages pre-trained language models like BERT to evaluate text generation tasks. Unlike traditional metrics like BLEU and ROUGE, which rely on exact word matches, BERTScore assesses the semantic similarity between the generated text and the reference text. Usually, it gives a fine-grained measure of semantic similarity based on cosine similarity
    \item \textbf{Semantic Accuracy (\underline{SA})} \cite{li2020explicit, segonne2023definition} is a measure of how faithfully the generated definition reflects the accepted understanding of the word's meaning, judged against a reference definition. BLEU, ROUGE-L, and BERTScore can provide some indication of similarity to a reference definition, and if the reference definition is accurate, a high score might suggest some level of semantic accuracy. However, they are easily fooled by surface-level similarities and don't explicitly assess the validity of the generated definition's meaning. Therefore, they are insufficient and unreliable as direct measures of semantic accuracy. Therefore, we utilize LLM-based evaluation, detailed in Appendix \ref{thise}, which is better suited for evaluating semantic accuracy.
    \item \textbf{Semantic Completeness (\underline{SC})} \cite{li2020explicit} (or called Factuality \cite{segonne2023definition}) refers to a definition encompassing all and only the relevant aspects of a word’s meaning. As illustrated in \cite{li2020explicit}, accurately defining "captain" in the context "The captain gave the order to abandon the ship" requires knowing that (1) a captain is a person, (2) a captain works on a ship, and (3) a captain is typically responsible for the ship. To assess SC, we employ a LLM-based evaluation, the specifics of which are detailed in Appendix \ref{thise}.
\end{itemize}

\section{Additional Analysis on \ours}
\label{additional}
Our primary objective is not to introduce a universally superior method that achieves state-of-the-art performance across all metrics and LLM backbones. Instead, this paper presents a benchmark study to investigate the fundamental question of whether LLMs can effectively understand internet buzzwords. While our proposed method \ours~demonstrates some performance improvements over existing approaches (cf. Section \ref{overll}), we include it within our benchmark analysis to critically assess the limitations of current methods, including our own (cf. Section \ref{inp}).

Beyond the benchmark analysis, this section aims to further explore the unique characteristics of our method \ours, offering readers a new perspective to understand its underlying principles. In particular, we have the following observations.

\noindent\textbf{\textit{Why does \ours~exhibit inferior performance under smaller LLMs?} -- This stems from LLM's difficulty in understanding and applying the illustrative aspects that guide buzzword definition generation}. As evidenced by Tables \ref{tab:benchmark} and \ref{tab:compare}, no single method consistently achieves the best performance across all metrics and LLM backbones. Regarding our proposed method \ours, it demonstrates superior performance compared to FOCUS when utilizing larger LLM backbones, while achieving comparable performance to FOCUS with smaller LLMs. Given that \ours~leverages key skills of child learning and codifies them into illustrative aspects to guide LLM-driven buzzword definition generation, we hypothesize that its less-than-optimal performance on smaller LLMs may stem from their difficulty in understanding these illustrative aspects and, consequently, in generating accurate definitions. To verify this hypothesis, we manually checked 100 buzzwords and the corresponding definitions generated by both Qwen2-7b-based \ours~and GPT-4o-based \ours. For each LLM and each buzzword, two human evaluators independently assessed whether each aspect-guided definition conformed to the specific aspect's requirements. As shown in Table \ref{apap}, we found that, on average, Qwen2-7b generates a definition consistent with the specific aspect requirements in 41.8\% of cases, while GPT-4o achieved a success rate of 57.2\%. Furthermore, we found that the aspects, PS and CA, are particularly challenging for both LLM.

\begin{table}[]
\centering
\resizebox{0.5\textwidth}{!}{%
\begin{tabular}{lll}
\toprule
\textbf{Aspects}               & \textbf{GPT-4o-based \ours} & \textbf{Qwen2-7b-based \ours} \\ \midrule
Intention Understanding, IU    & 22.97\%               & 22.58\%                 \\
Concept Association, CA        & 18.92\%               & 22.58\%                 \\
Language Structure, LS         & 18.92\%               & 12.90\%                 \\
Social Cue Interpretation, SCI & 10.81\%               & 11.29\%                 \\
Word Context, WC               & 18.92\%               & 17.74\%                 \\
Pronunciation and Spelling, PS & 9.46\%                & 12.90\%       \\ \bottomrule
\end{tabular}%
}
\caption{The average selection rate for each of the six aspects that most effectively guide the final definition.}
\label{apap2}
\end{table}

\noindent\textbf{\textit{Why does adding more than three aspects result in minimal performance gains?} -- Understanding a word may only require a subset of key aspects}. While Figure \ref{fig:aspect} demonstrates that the performance of our method does improve with an increasing number of aspects, the rate of improvement diminishes as more aspects are added. We hypothesize that this trend arises because, for both LLMs and humans, understanding a word may not require the consideration of all possible aspects; perhaps a subset of key aspects is sufficient. Furthermore, this subset may vary depending on the word. This would explain why providing more aspects to the LLM leads to performance gains—the newly added aspects may offer enhanced understanding for a subset of words, even if not for all. To validate this hypothesis, we conducted further experiments, manually checking 100 buzzwords and their corresponding definitions generated by both Qwen2-7b-based RESS and GPT-4o-based RESS. For each buzzword and LLM, two individuals independently selected up to three of the six aspect-guided definitions that were most helpful in guiding the final definition generated by each model (e.g., definitions that were semantically closest, provided usage scenarios, revealed original meanings, or presented additional information like figurative meanings). Table \ref{apap2} records the averaged selection percentages for each of the six aspects. As these results illustrate, not all aspects contribute equally to the generation of a final definition. Some aspects appear to be more beneficial for understanding specific buzzwords, leading to less significant overall performance improvements on a dataset-wide scale. This finding supports our previous hypothesis, as it indicates that: 1) using more aspects can be beneficial and 2) some aspects may be more helpful for understanding a specific subset of buzzwords, thereby contributing to overall dataset performance without all aspects being equally important.

\begin{table}[]
\centering
\resizebox{0.5\textwidth}{!}{%
\begin{tabular}{lll}
\toprule
\textbf{Aspects}               & \textbf{GPT-4o-based \ours} & \textbf{Qwen2-7b-based \ours} \\ \midrule
Intention Understanding, IU    & 98.5                  & 96                      \\
Concept Association, CA        & 28                    & 29                      \\
Language Structure, LS         & 64                    & 7.5                     \\
Social Cue Interpretation, SCI & 46                    & 38                      \\
Word Context, WC               & 99                    & 78.5                    \\
Pronunciation and Spelling, PS & 7.5                   & 1.5                     \\ \midrule
Avg.                           & 57.2                  & 41.8                   \\ \bottomrule
\end{tabular}%
}
\caption{Accuracy (\%) of human evaluation in judging whether aspect-guided definitions conform to specific aspect requirements.}
\label{apap}
\end{table}

\noindent\textbf{\textit{Why do \ours~and FOCUS perform poorly under conventional metrics compared to other methods?} -- \ours~and FOCUS generate free-form, lengthy definitions with elaborations that are penalized by the n-gram matching and similarity calculations inherent in these metrics.}. The results presented in Table \ref{tab:benchmark} and Table \ref{tab:compare} indicate that neither FOCUS nor our proposed RESS method consistently outperforms the DP and CoT baselines according to conventional metrics. This discrepancy arises primarily because FOCUS and RESS tend to generate free-form and long buzzword definitions, often including elaborations on the figurative meaning and connotations of the target buzzword. Given that conventional metrics such as BLEU and R-L rely on n-gram word matching, they inherently penalize free-form, lengthy responses that do not precisely replicate the vocabulary of the gold reference (i.e., the ground-truth definitions). Consequently, generating longer definitions without using the exact words from the ground truth can lead to lower BLEU and R-L scores. Furthermore, the lower BScore observed for FOCUS and RESS compared to DP and CoT can be attributed to the additional explanatory content regarding the extended figurative meanings. This supplementary information may introduce semantic noise during BERT's similarity calculations. These inherent limitations of conventional metrics serve as a key motivation for incorporating LLM-based evaluation. As demonstrated in the LLM evaluation results of Table \ref{tab:benchmark} and Table \ref{tab:compare}, FOCUS and RESS are indeed shown to be superior to DP and CoT, a finding that aligns with our human evaluation results. To illustrate this point further, we provide two case studies in Appendix \ref{case}, offering a more intuitive demonstration of how the definitions generated by FOCUS and RESS include elaborations on the extended meanings, resulting in significantly longer outputs compared to other methods.

\section{Methods for Dictionary Example Selection}
\label{select}
While high-quality examples effectively illustrate a word's meaning and typical usage, identifying or creating such examples can be a laborious and costly endeavor \cite{stankovic2019sasa, de2009extracting}. While methods for automatic example selection have been proposed, they frequently rely on pre-existing word definitions as supervised signals to train a model and subsequently locate suitable examples \cite{kathuria2012word, shinnou-sasaki-2008-division, tolmachev2022automatic, benedetti2024automatically}, a strategy that is not applicable in our case since the definition of the buzzword is initially unknown (See Section \ref{ugc}). Instead, a limited number of studies have proposed rule-based methods without relying on word definitions. These methods, for example, GDEX \cite{kilgarriff2008gdex}, prioritize readability and informativeness, measured by, for example, sentence length, word frequencies, and syntactic information \cite{pilan-etal-2016-candidate, didakowski2012automatic, stankovic2019sasa}. In this paper, we also present a novel example selection method that bypasses the need for meticulously constructed rules.

\section{Case Studies}
\label{case}
This section provides case studies for better understanding the performance of different methods. Table \ref{tab:bcase} showcases the case of buzzword '\begin{CJK}{UTF8}{gbsn}窝囊费\end{CJK}'. According to the case study, \ours~generates more nuanced definitions, capturing not only the concept of hard work itself but also the frustration and resignation stemming from the contrast between hard work and meager compensation. Moreover, Table \ref{tab:bcase2} shows another example of buzzword '\begin{CJK}{UTF8}{gbsn}梁静茹给的勇气\end{CJK}'. In this case, both the DP and CoT fail to deliver accurate definition. Additionally, the definitions generated by FOCUS erroneously include the sense of "willingness to proceed".

\begin{table}[t]
\centering
\resizebox{0.5\textwidth}{!}{%
\begin{tabular}{l}
\toprule
\textbf{Internet Buzzword} \\ \midrule
\begin{tabular}[c]{@{}l@{}}\begin{CJK}{UTF8}{gbsn}梁静茹给的勇气\end{CJK}  (The courage given by Fish Leong)\end{tabular} \\ \midrule
\textbf{Description from Online Source} \\ \midrule
\begin{tabular}[c]{@{}l@{}}\begin{CJK}{UTF8}{gbsn}梁静茹给的勇气，多是疑问句，\end{CJK}\\ \begin{CJK}{UTF8}{gbsn}是梁静茹给你的勇气吗？\end{CJK}\\ \begin{CJK}{UTF8}{gbsn}意思是你哪来的勇气说出这句话或者干出这种事，是在嘲讽。\end{CJK}\\ 
(The phrase "The courage given by Fish Leong" is often used sarcastically\\ as a rhetorical question, implying, "Where did you get the nerve to\\ say/do that?" Usually, it's a way of mocking someone's audacity.)\end{tabular} \\ \midrule
\textbf{Definition} \\ \midrule
\begin{tabular}[c]{@{}l@{}}\begin{CJK}{UTF8}{gbsn}嘲讽某人哪来的勇气说出这句话或做出这件事\end{CJK}\\ (To sarcastically mock someone for their audacious behavior)\end{tabular} \\ \midrule

\textbf{Examples (i.e., UGC). We only show one example here to save space} \\ \midrule
\begin{tabular}[c]{@{}l@{}}\begin{CJK}{UTF8}{gbsn}oppo用脚做pro，直接一整个退订，梁静茹给的勇气吧，\end{CJK}\\\begin{CJK}{UTF8}{gbsn}短焦砍主摄还敢定价这么高，我的小米继续坚挺\end{CJK}\\ 
(Oppo's Pro model is a complete letdown; I've canceled my pre-order. \\
They must have gotten their courage from Fish Leong to cut the main camera\\
 on a short-focus lens and still charge such a high price. \\My Xiaomi phone will continue to serve me well.)\end{tabular} \\ \midrule

\textbf{Definitions of DP} \\ \midrule
\begin{tabular}[c]{@{}l@{}}\begin{CJK}{UTF8}{gbsn}一种源自梁静茹音乐作品影响下的勇气，\end{CJK}\\\begin{CJK}{UTF8}{gbsn}鼓励人们在面对挑战或困难时勇敢做出决策和行动的心态。\end{CJK}\\ 
(A courage inspired by the music of Fish Leong, \\ 
empowering individuals to make bold decisions and \\take action in the face of challenges and adversity.)\end{tabular} \\ \midrule

\textbf{Definitions of CoT} \\ \midrule
\begin{tabular}[c]{@{}l@{}}\begin{CJK}{UTF8}{gbsn}指因受到梁静茹的音乐或精神影响而产生的勇气与决心，\end{CJK}\\\begin{CJK}{UTF8}{gbsn}常用于形容在面对困难或挑战时，\end{CJK}\\\begin{CJK}{UTF8}{gbsn}勇敢做出决定或采取行动的状态。\end{CJK}\\ 
(It refers to the courage and resolve inspired by the music or \\ spirit of singer Fish Leong, often used to describe the state \\of bravely making decisions or taking action in the face of difficulty or challenge.)\end{tabular} \\ \midrule

\textbf{Definitions of FOCUS} \\ \midrule
\begin{tabular}[c]{@{}l@{}}\begin{CJK}{UTF8}{gbsn}梁静茹给的勇气表示一种因冲动或勇敢而做出的决定或行动，\end{CJK}\\\begin{CJK}{UTF8}{gbsn}常带有调侃和自嘲的意味。\end{CJK}\\\begin{CJK}{UTF8}{gbsn}它经常被用于描述在面对困难、挑战或不确定性时，\end{CJK}\\\begin{CJK}{UTF8}{gbsn}做出大胆选择或尝试的情景。这个短语通常反映出一种无畏的态度，\end{CJK}\\\begin{CJK}{UTF8}{gbsn}可能隐含着对所做决定的无奈或讽刺，\end{CJK}\\\begin{CJK}{UTF8}{gbsn}表示虽然心中有一种不安，但仍然选择去尝试。\end{CJK}\\ 
(It refers to a decision or action driven by impulsiveness or bravery, \\often with a self-deprecating or ironic undertone. It's frequently used to \\describe bold choices or attempts in the face of difficulty, \\challenges, or uncertainty. The phrase typically reflects\\ a fearless attitude, possibly tinged with resignation or sarcasm regarding the decision,\\ suggesting a sense of apprehension alongside the willingness to proceed.)\end{tabular} \\ \midrule

\textbf{Definitions of \ours} \\ \midrule
\begin{tabular}[c]{@{}l@{}}\begin{CJK}{UTF8}{gbsn}梁静茹给的勇气是一种调侃的表达，用来形容某人做出大胆或不理智的行为，\end{CJK}\\\begin{CJK}{UTF8}{gbsn}通常带有讽刺或自嘲的意味，暗示该行为的勇气似乎是凭空而来，\end{CJK}\\\begin{CJK}{UTF8}{gbsn}不切实际。这种表达源自歌手梁静茹的歌曲《勇气》，\end{CJK}\\\begin{CJK}{UTF8}{gbsn}常用于描述一些看似勇敢但并未经过深思熟虑的决定或行动\end{CJK}\\ 
(It is an ironic expression used to describe bold or \\reckless behavior, often with a sarcastic or self-deprecating tone. \\It implies that the courage seems unfounded and unrealistic, \\referencing the singer Fish Leong's song of the same name ("Courage").\\ This expression is commonly used to describe decisions or actions \\that appear brave but lack careful consideration.)\end{tabular} \\  \bottomrule
\end{tabular}
}
\caption{Case of buzzword '\begin{CJK}{UTF8}{gbsn}梁静茹给的勇气\end{CJK}' and generated definitions by different methods. Here, we also provide its English translation for better understanding.}
\setlength{\abovecaptionskip}{2pt}   
\setlength{\belowcaptionskip}{2pt}
\label{tab:bcase2}
\end{table}

\begin{table}[t]
\centering
\resizebox{0.5\textwidth}{!}{%
\begin{tabular}{l}
\toprule
\textbf{Internet Buzzword} \\ \midrule
\begin{tabular}[c]{@{}l@{}}\begin{CJK}{UTF8}{gbsn}窝囊费\end{CJK}  (gutlessness fee)\end{tabular} \\ \midrule
\textbf{Description from Online Source} \\ \midrule
\begin{tabular}[c]{@{}l@{}}\begin{CJK}{UTF8}{gbsn}指那些在职场上辛苦耕耘、忍气吞声，最终换来的微薄薪水。\end{CJK}\\ (Refers to those who work hard and endure hardships in the workplace, \\ultimately earning meager salaries.)\end{tabular} \\ \midrule
\textbf{Definition} \\ \midrule
\begin{tabular}[c]{@{}l@{}}\begin{CJK}{UTF8}{gbsn}指工作辛苦劳累、受委屈，得到的回报却很少。\end{CJK}\\ (Despite significant effort and hardship endured, the salary received is minimal)\end{tabular} \\ \midrule

\textbf{Examples (i.e., UGC). We only show one example here to save space} \\ \midrule
\begin{tabular}[c]{@{}l@{}}\begin{CJK}{UTF8}{gbsn}早知道就考体制内了，也不至于月月赚窝囊费\end{CJK}\\ (I should have gone for a
government job. \\At least I wouldn't be earning gutlessness fee every month.)\end{tabular} \\ \midrule

\textbf{Definitions of DP} \\ \midrule
\begin{tabular}[c]{@{}l@{}}\begin{CJK}{UTF8}{gbsn}形容微薄、无足轻重的收入或工资，通常带有不满或无奈的情绪\end{CJK}\\ (Describing meager and insignificant income or salary, \\usually accompanied by dissatisfaction or helplessness)\end{tabular} \\ \midrule

\textbf{Definitions of CoT} \\ \midrule
\begin{tabular}[c]{@{}l@{}}\begin{CJK}{UTF8}{gbsn}指微薄的工资或收入，通常带有无奈、无力感，反映出工作的辛苦\end{CJK}\\ \begin{CJK}{UTF8}{gbsn}与经济压力，常用于表达对现状的不满或无奈\end{CJK}\\ (Refers to meager wages or income, usually with a sense of \\helplessness and powerlessness, reflecting the hardship and \\economic pressure of work, often used to express dissatisfaction \\or helplessness towards the current situation.)\end{tabular} \\ \midrule

\textbf{Definitions of FOCUS} \\ \midrule
\begin{tabular}[c]{@{}l@{}}\begin{CJK}{UTF8}{gbsn}窝囊费表示一种微薄且让人感到无奈的工资或报酬。\end{CJK}\\ \begin{CJK}{UTF8}{gbsn}它经常被用于形容工作带来的失望和对生活状态的自嘲。\end{CJK}\\ \begin{CJK}{UTF8}{gbsn}这个短语常常隐含着对工作的无奈与对生活现状的不满，\end{CJK}\\\begin{CJK}{UTF8}{gbsn}表达出人们在日常工作中所感受到的疲惫与无力感。\end{CJK}\\ (Gutlessness fee refers to meager and disheartening pay or compensation. \\It's often used to express the disappointment stemming from one's work \\and a self-deprecating attitude towards one's life circumstances.\\ The term subtly conveys feelings of helplessness about work and \\dissatisfaction with one's current life situation, reflecting the weariness \\and powerlessness many experience in their daily work lives.)\end{tabular} \\ \midrule

\textbf{Definitions of \ours} \\ \midrule
\begin{tabular}[c]{@{}l@{}}\begin{CJK}{UTF8}{gbsn}窝囊费是指一种微薄或低质量的收入，通常被用来形容工资、\end{CJK}\\
\begin{CJK}{UTF8}{gbsn}报酬或补贴，带有无奈、轻蔑和自嘲的情感色彩，\end{CJK}\\
\begin{CJK}{UTF8}{gbsn}反映出劳动者对工作状况和生活质量的不满与失望。\end{CJK}\\
\begin{CJK}{UTF8}{gbsn}这个词语常常在语境中体现出对收入微薄与辛苦工作的无奈，\end{CJK}\\
\begin{CJK}{UTF8}{gbsn}表达人们对自身价值的反思和对现实的不满\end{CJK}\\ \begin{CJK}{UTF8}{gbsn}与经济压力，常用于表达对现状的不满或无奈\end{CJK}\\ (Gutlessness fee refers to meager or low-quality income, \\typically used to describe wages, remuneration, or subsidies.\\ It carries a strong emotional undertone of resignation, disdain, \\and self-deprecation, reflecting the worker's dissatisfaction \\and disappointment with their work conditions and quality of life. \\The term often highlights the irony of low pay in the face of \\arduous work, expressing a sense of self-reflection on \\one's value and discontent with the current circumstances.)\end{tabular} 
\\ \bottomrule
\end{tabular}
}
\caption{Case of buzzword '\begin{CJK}{UTF8}{gbsn}窝囊费\end{CJK}' and generated definitions by different methods. Here, we also provide its English translation for better understanding.}
\setlength{\abovecaptionskip}{2pt}   
\setlength{\belowcaptionskip}{2pt}
\label{tab:bcase}
\end{table}

\begin{table*}[!t]
    \centering
    \resizebox{0.9\textwidth}{!}{%
    \begin{tabular}{p{0.97\textwidth}}
    \toprule
    \textbf{\textit{Prompt for Aspect-specific definition generation}}     \\
    \midrule
    \begin{CJK}{UTF8}{gbsn}
    根据以下所有[例句]，分析词语\end{CJK}\begin{CJK}{UTF8}{gbsn}[BUZZWORD]\end{CJK}\begin{CJK}{UTF8}{gbsn}的含义，将其总结成一句通顺且易理解的定义，并简要解释原因。
    \end{CJK}\\
    \begin{CJK}{UTF8}{gbsn}
        注意：
    \end{CJK}\\
    \begin{CJK}{UTF8}{gbsn}
        1. 用中文回答
    \end{CJK}\\
    \begin{CJK}{UTF8}{gbsn}
        2. 你需要从\end{CJK}\begin{CJK}{UTF8}{gbsn}[INPUT\_ASPECT]\end{CJK}\begin{CJK}{UTF8}{gbsn}角度一步一步地思考这个词语的定义，这意味着去理解\end{CJK}\begin{CJK}{UTF8}{gbsn}[INPUT\_ASPECT\_EXPLANATION]\end{CJK}\begin{CJK}{UTF8}{gbsn}
    \end{CJK}\\
    \begin{CJK}{UTF8}{gbsn}
        3. 在观察用法示例时，要彻底解释上下文，以推断短语的微妙含义。将你的推理分解为循序渐进的逻辑，以达成全面的理解
    \end{CJK}\\
    \begin{CJK}{UTF8}{gbsn}
        4. 你不能过度解读这个词
    \end{CJK}\\
    \begin{CJK}{UTF8}{gbsn}
        5. 以Json形式返回结果：\{"词语": "\end{CJK}\begin{CJK}{UTF8}{gbsn}[BUZZWORD]\end{CJK}\begin{CJK}{UTF8}{gbsn}", "定义": STRING, "原因"：STRING\}
    \end{CJK}\\
    \begin{CJK}{UTF8}{gbsn}
        [生成示例]: \end{CJK}\begin{CJK}{UTF8}{gbsn}[EXAMPLES]\end{CJK}\begin{CJK}{UTF8}{gbsn}
    \end{CJK}\\
    \begin{CJK}{UTF8}{gbsn}
        ==================
    \end{CJK}\\
    \begin{CJK}{UTF8}{gbsn}
        [例句]: 
    \end{CJK}[UGC\_SENTENCES]\\\\
    Based on all the following [Example Sentences], analyze the meaning of the word [BUZZWORD], summarize it into a coherent and easy-to-understand definition, and briefly explain the reason.
\\
be careful:\\
1. Answer in Chinese\\
2. You need to think step by step about the definition of this word from the perspective of [INPUT\_ASPECT], which means understanding [INPUT\_ASPECT\_EXPLANATION]\\
3. When observing usage examples, thoroughly explain the context to infer the subtle meaning of the phrase. Break down your reasoning into progressive logic to achieve a comprehensive understanding\\
4. You cannot overinterpret this word\\
5. Return the result in JSON format: \{"Word": "[BUZZWORD]", "Definition": STRING, "Reason": STRING\}\\

[Example of Generation]: [EXAMPLES]\\
\begin{CJK}{UTF8}{gbsn}
==================
\end{CJK}\\
Example Sentences: [UGC\_SENTENCES] \\
    \bottomrule
    \end{tabular}
    }
    \caption{Prompt for \ours~and its corresponding translation: Part I.}
    \label{prompt:aspects}
\end{table*}

\begin{table*}[!t]
    \centering
    \resizebox{0.9\textwidth}{!}{%
    \begin{tabular}{p{0.97\textwidth}}
    \toprule
    \textbf{\textit{Prompt for ensembling aspect-specific definition candidates}}     \\
    \midrule
    \begin{CJK}{UTF8}{gbsn}
    根据以下所有[例句]，分析词语\end{CJK}\begin{CJK}{UTF8}{gbsn}[BUZZWORD]\end{CJK}\begin{CJK}{UTF8}{gbsn}的含义，总结该词的[参考定义]成通顺且易理解的定义，包括但不限于本义、引申义和用法等等，并简要解释原因。
    \end{CJK}\\
    \begin{CJK}{UTF8}{gbsn}
    注意：
    \end{CJK}\\
    \begin{CJK}{UTF8}{gbsn}
    1. 用中文回答
    \end{CJK}\\
    \begin{CJK}{UTF8}{gbsn}
    2. 你需要根据[例句]一步一步分析该词[参考定义]的重要性，不是所有的[参考定义]都是有价值的。
    \end{CJK}\\
    \begin{CJK}{UTF8}{gbsn}
    3. 在分析时，要结合[例句]和[参考定义]，以推断[参考定义]的微妙含义，以达成全面的理解。
    \end{CJK}\\
    \begin{CJK}{UTF8}{gbsn}
    4. 以Json形式返回结果：\{"词语": "\end{CJK}\begin{CJK}{UTF8}{gbsn}[BUZZWORD]\end{CJK}\begin{CJK}{UTF8}{gbsn}", "定义": STRING, "原因"：STRING\}
    \end{CJK}\\
    \begin{CJK}{UTF8}{gbsn}
    [生成示例]: \end{CJK}[EXAMPLES]\\
    \begin{CJK}{UTF8}{gbsn}
    ==================    
    \end{CJK}\\
    \begin{CJK}{UTF8}{gbsn}
    [参考定义]:
    \end{CJK}[CANDIDATE\_DEFINITION]\\
    \begin{CJK}{UTF8}{gbsn}
    [例句]:
    \end{CJK} [UGC\_SENTENCES]
    \\\\
    Based on all the following [Example Sentences], analyze the meaning of the word [BUZZWORD], summarize its [Reference Definitions] into a coherent and easy-to-understand definition, including but not limited to its original meaning, extended meaning, usage, etc., and briefly explain the reasons.\\
be careful:\\
1. Answer in Chinese\\
2. You need to analyze the importance of the word [reference definition] step by step based on [Example Sentences], not all [Reference Definitions] are valuable.\\
3. When analyzing, it is necessary to combine [example sentence] and [reference definition] to infer the subtle meaning of [reference definition] in order to achieve a comprehensive understanding.\\
4. Return the result in JSON format: \{"Word": "[BUZZWORD]", "Definition": STRING, "Reason": STRING\}
\\
Example of Generation: [EXAMPLES]\\
==================\\
Reference Definitions: [CANDIDATE\_DEFINITION]\\
Example Sentences: [UGC\_SENTENCES]\\

    \bottomrule
    \end{tabular}
    }
    \caption{Prompt for \ours~and its corresponding translation: Part II.}
    \label{prompt:emsenble}
\end{table*}

\begin{table*}[!t]
    \centering
    \resizebox{0.9\textwidth}{!}{%
    \begin{tabular}{p{0.97\textwidth}}
    \toprule
    \textbf{\textit{Prompt for DP$_{\text{-w/o UGC}}$}}     \\
    \midrule
    \begin{CJK}{UTF8}{gbsn}
    给出以下互联网流行词或短语的定义。\end{CJK}\\
     \begin{CJK}{UTF8}{gbsn}
    注意，
    \end{CJK}\\
     \begin{CJK}{UTF8}{gbsn}
    1. 你给出的定义需要是简洁易懂的一句或多句话
    \end{CJK}\\
     \begin{CJK}{UTF8}{gbsn}
    2. 以json形式返回结果：\{'word': STRING, 'definition': STRING\}
    \end{CJK}\\
     \begin{CJK}{UTF8}{gbsn}
    词语：
    \end{CJK}[BUZZWORD]
   \\\\
   Return definitions of the following Internet buzzwords or phrases.\\
be careful,\\
1. The definition you provide needs to be concise and easy to understand\\
2. Return the result in JSON format: \{'word ': STRING,' definition ': STRING\}\\
    
Words: [BUZZWORD]\\
    \bottomrule
    \end{tabular}
    }
    \caption{Prompt for DP$_{\text{-w/o UGC}}$, which is also used in contamination-free evaluation experiments.}
    \label{key_prompt}
\end{table*}

\begin{table*}[!t]
    \centering
    \resizebox{0.9\textwidth}{!}{%
    \begin{tabular}{p{0.97\textwidth}}
    \toprule
    \textbf{\textit{Prompt for Aspect-specific definition generation}}     \\
    \midrule
    \begin{CJK}{UTF8}{gbsn}
    给定一个词语的【定义】和专家给出的【参考定义】，你需要从以下【评估角度和打分标准】，为这个【定义】的质量高低评分。\end{CJK}\\\begin{CJK}{UTF8}{gbsn}
    使用Json格式返回结果：\{"准确性": [INT, WHY], "细节完整性": [INT, WHY]\}\end{CJK}\\\\\begin{CJK}{UTF8}{gbsn}
    【定义】：[PREDICTED\_DEFINITION]\end{CJK}\\\begin{CJK}{UTF8}{gbsn}
    【参考定义】：[GROUND\_TRUTH\_DEFINITION]\end{CJK}\\\begin{CJK}{UTF8}{gbsn}
    【评估角度和打分标准】：\end{CJK}\\\begin{CJK}{UTF8}{gbsn}
    ===================\end{CJK}\\\begin{CJK}{UTF8}{gbsn}
    准确性：\end{CJK}\\\begin{CJK}{UTF8}{gbsn}
    1分：该定义与【参考定义】相比，严重偏离了词语的真实意义，或者包含大量错误信息。\end{CJK}\\\begin{CJK}{UTF8}{gbsn}
    2分：该定义与【参考定义】相比，有一定的偏差，但至少部分正确。\end{CJK}\\\begin{CJK}{UTF8}{gbsn}
    3分：该定义与【参考定义】相比，基本准确，但可能存在一些小错误或不完整的描述。\end{CJK}\\\begin{CJK}{UTF8}{gbsn}
    4分：该定义与【参考定义】相比，准确，能够清晰地传达词语的核心意义。\end{CJK}\\\begin{CJK}{UTF8}{gbsn}
    5分：该定义与【参考定义】相比，非常准确，全面反映了词语的意义，没有遗漏重要细节。\end{CJK}\\\\\begin{CJK}{UTF8}{gbsn}
    细节完整性：\end{CJK}\\\begin{CJK}{UTF8}{gbsn}
    1分：该定义与【参考定义】相比，遗漏了许多重要的细节。\end{CJK}\\\begin{CJK}{UTF8}{gbsn}
    2分：该定义与【参考定义】相比，遗漏了一些重要的细节，但整体还算完整。\end{CJK}\\\begin{CJK}{UTF8}{gbsn}
    3分：该定义与【参考定义】相比，包含大部分必要细节，但仍有改进空间。\end{CJK}\\\begin{CJK}{UTF8}{gbsn}
    4分：该定义与【参考定义】相比，包含了几乎所有必要的细节。\end{CJK}\\\begin{CJK}{UTF8}{gbsn}
    5分：该定义与【参考定义】相比，包含了所有必要的细节，没有遗漏。\end{CJK}\\\begin{CJK}{UTF8}{gbsn}
    \end{CJK}\\
    \begin{CJK}{UTF8}{gbsn}===================\end{CJK}\\\\

    Given a definition of a word and its Reference Definition provided by experts, you need to rate the quality of the definition from the following evaluation perspectives and scoring criteria. \\
    Return result in JSON format: \{'SA': [INT, WHY], 'SC': [INT, WHY]\}\\
    \\
    Definition: [PREDICTED\_DEFINITION]\\
    Reference Definition: [GROUND\_TRUTH\_DEFINITION]\\
    Evaluation perspective and scoring criteria:\\
    ===================\\
    Semantic Accuracy (SA):\\
    1 point: Compared with the Reference Definition, this definition deviates significantly from the true meaning of the word or contains a large amount of erroneous information.\\
    2 points: This definition has some deviation compared to the Reference Definition, but at least partially correct.\\
    3 points: Compared with the Reference Definition, this definition is generally accurate, but there may be some minor errors or incomplete descriptions.\\
    4 points: Compared with the Reference Definition, this definition is accurate and can clearly convey the core meaning of the word.\\
    5 points: Compared with the Reference Definition, this definition is very accurate and fully reflects the meaning of the words, without missing any important details.\\\\
        
    Semantic Completeness (SC):\\
    1 point: Compared with the Reference Definition, this definition misses many important details.\\
    2 points: Compared with the Reference Definition, this definition has omitted some important details, but overall it is relatively complete.\\
    3 points: Compared with the Reference Definition, this definition contains most of the necessary details, but there is still room for improvement.\\
    4 points: Compared to the Reference Definition, this definition contains almost all necessary details.\\
    5 points: Compared with the Reference Definition, this definition includes all necessary details without omission.\\
    ===================\\

    \bottomrule
    \end{tabular}
    }
    \caption{Prompt for GPT-4o evaluator.}
    \label{evaluation_prompt}
\end{table*}

\end{document}